\documentclass{article}

\PassOptionsToPackage{numbers, compress}{natbib}
\usepackage{makecell}

\usepackage{float}

\usepackage[preprint]{neurips_2024}

\usepackage[utf8]{inputenc} 
\usepackage[T1]{fontenc}    
\usepackage{hyperref}       
\usepackage{url}            
\usepackage{booktabs}       
\usepackage{amsfonts}       
\usepackage{nicefrac}       
\usepackage{microtype}      
\usepackage{xcolor}   
\usepackage{amsmath}
\usepackage{amssymb}
\usepackage{graphicx}
\usepackage{hyperref}
\usepackage{amsmath}
\usepackage{graphicx}
\usepackage{hyperref}
\usepackage{comment}
\usepackage[moderate]{savetrees}

\title{Modeling Language as a Sequence of Thoughts}

\author{
  Nasim Borazjanizadeh\thanks{Authors contributed equally to this work.} \\
  Stanford University \\
  \texttt{nasimb@stanford.edu} \\
  \And
  James L. McClelland\footnotemark[1] \\
  Stanford University \\
  \texttt{jlmcc@stanford.edu}
}

\begin{document}

\maketitle
\vspace{-0.3cm}

\begin{abstract}
\vspace{-0.08cm}
Transformer language models can generate strikingly natural text by modeling language as a sequence of tokens. Yet, by relying primarily on surface-level co-occurrence statistics, they fail to form globally consistent latent representations of entities and events, lack of which contributes to poor relational generalization (reversal curse), contextualization errors, and data inefficiency. On the other hand, cognitive science shows that human comprehension involves converting the input linguistic stream into compact, event-like representations that persist in memory while verbatim form is short-lived. Motivated by these cognitive findings, we introduce the \textbf{Thought Gestalt (TG) Model}, a recurrent transformer that models language at two levels of abstraction—tokens and sentence-level "thought" states. TG generates the tokens of one sentence at a time while cross-attending to a working memory of prior sentence representations. In TG, token and sentence representations are generated using a shared stack of transformer blocks and trained with a single objective, the next-token prediction loss: by retaining the computation graph of sentence representations written to the working memory, gradients from future token losses flow backward through cross-attention to optimize the parameters generating earlier sentence vectors. In scaling experiments, TG consistently improves data and parameter efficiency compared to matched GPT-2 runs, among other baselines, with scaling fits indicating GPT-2 requires \textasciitilde5--8\% more data and \textasciitilde33--42\% more parameters to match TG's test loss. TG also reduces errors in relational-direction generalization on a father–son reversal curse probe.
\vspace{-0.1cm}
\end{abstract}

\section{Introduction}
\vspace{-0.1cm}

Prior work in cognitive science suggests that, in humans, language functions as a serial code for communicating underlying thoughts, rather than the constitutive medium of thought itself \citep{shank1972conceptual,mcclelland2020placing, fedorenko2024language,pinker1990natural}. On this view, comprehension involves decoding a linguistic stream to construct a situation model—a mental representation that encodes the temporal sequences, causal relations, and entities of the described event \citep{zwaan1998situation,glenberg1987mental}. Situation models support memory and later inference and are characterized as high-level conceptual representations integrated with prior knowledge rather than representations of the surface form of text \citep{zwaan1998situation,mcclelland2020placing}.

In contrast, modern Large Language Models (LLMs) learn by modeling language as a sequence of tokens and optimizing next-token prediction \citep{openai2023gpt4,deepseekv3,llama3_2024}. Although this has yielded models with remarkable fluency, a token-centric training signal can encourage brittle heuristics that capture surface-level statistical patterns rather than underlying concepts, leading to failures in generalization and compositional tasks \citep{bender2020climbing,dziri2024faithandfate,mirzadeh2024gsmsymbolic}. The reversal curse is a concrete example: models trained on one relational direction (e.g., ``A is B'') often fail to generalize to the inverse (``B is A''), treating the two directions of a relational fact as distinct patterns rather than a unified semantic encoding \citep{berglund2024reversal,lin2024delving}. While \citet{berglund2024reversal} note that the reversal curse is not observed for in-context learning in their evaluated models (e.g., GPT-4), in \S\ref{sec:results-reversal} we show that a substantial in-context directional asymmetry exists in smaller models at the scale of GPT-2 when a relation is queried in the opposite direction of the prompt.

Another shortcoming of standard transformer models is the contextualization errors identified by \citet{lepori2024racingthoughts}, where in lower transformer layers, later tokens attend to earlier token representations before they have been fully contextualized to resolve ambiguities. Finally, many modern models are trained on trillions of tokens \citep{deepseekv3}, exceeding estimates of children’s language exposure by orders of magnitude (on the order of tens of millions of words \citep{hart_risley_1995_wordgap}). These gaps motivate architectures that learn and reuse latent representations at a higher level of abstraction than tokens, i.e., models that organize information into coherent \emph{gestalts}—holistic representations whose properties are not reducible to the sum of their parts \citep{werthiemer1938gestalt}—that capture the underlying concepts and relations expressed by text \citep{mcclelland2020placing,glenberg1987mental}.

\begin{figure}[t]
\centering
\includegraphics[width=0.7\linewidth]{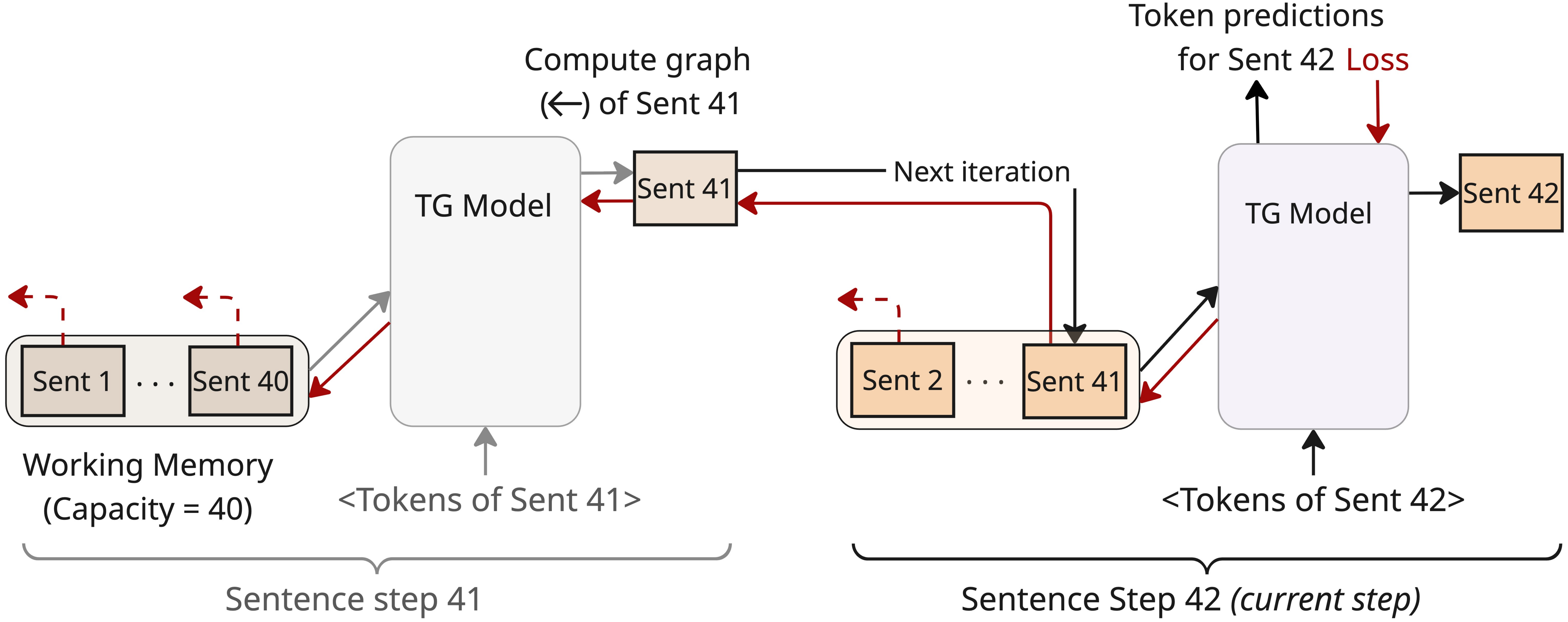}


\caption{\small \textbf{Forward/backward pass in TG.} Each sentence step produces next-token predictions and a sentence vector, which is appended to a fixed-capacity memory without detaching its computation graph (removing the oldest memory entry if full). Next-token loss gradients flow back through memory to optimize parameters that produced earlier sentence representations (see Fig.~\ref{fig:tg_grad} for a more detailed gradient-flow view).}

\vspace{-0.5cm}
\label{fig:tg-flow}
\end{figure}

In this work, we introduce the \emph{Thought Gestalt} (TG) model, a recurrent transformer architecture designed to model language at two levels of abstraction: tokens and sentence-level thoughts. TG processes one sentence at a time (a "sentence step"), maintaining  token-level information of only the current sentence, while context is maintained as a \emph{working memory} of prior sentence representations: holistic representations, or "gestalts", each compressing an entire sentence into a vector  (Figure~\ref{fig:tg-flow}). This design is motivated by cognitive evidence that humans segment continuous streams of text into discrete events to organize memory: the verbal form is short-lived and confined to a narrow span, while earlier content is retained as stable, high-level representations of events and relations \citep{radvansky2011event,jarvella1971syntactic,zwaan2016situation}. TG is also inspired by the Sentence Gestalt model \citep{stjohn1990learning}, which incrementally maps a word sequence onto a single event representation from which role--filler structure can be decoded. TG similarly uses sentence boundaries as a structural proxy for thought boundaries. While not a one-to-one mapping, sentence boundaries serve as natural cognitive junctures where background information is integrated and concepts are updated \citep{guzman2000maintaining,magliano2007situation}. Learning sentence representations is thus a first step toward building generative systems that can learn situation models and latent thought representations.

Architecturally, TG interleaves self-attention over tokens of the current sentence with cross-attention to a fixed-capacity working memory of prior sentence representations, where tokens attend causally within the current sentence and can reach prior sentences only via sentence-gestalt vectors (similar to gisting \citep{mu2023gisting}). Each sentence representation is built from the contextualized hidden state at the end-of-sentence (\texttt{<EOS>}) token position (Figure~\ref{fig:tg-arch}). After each sentence step, the new sentence representation is appended to memory and the oldest entry is evicted when the memory is full. Crucially, TG retains the computation graph of sentence representations when writing to the working memory, allowing next-token prediction losses from later sentences to backpropagate through cross-attention into the parameters that produced earlier sentence gestalts (see Figure~\ref{fig:tg-flow}). Because sentence gestalts are contextualized summaries of semantically coherent chunks of prior content, they are natural candidates for caching and retrieval at inference time to enable continual learning (as the Memorizing transformer \citep{wu2022memorizing}); here we focus on their role as a differentiable context for pretraining.

Our design of sentence representations learned end-to-end under the standard next-token objective distinguishes TG from recent architectures that learn sentence embeddings via auxiliary objectives, such as next-sentence prediction, token reconstruction, sentence ordering, or contrastive alignment, on top of a language-modeling loss \citep{devlin2019bert,duquenne2023sonar,lan2019albert,gao2021simcse,kashyap2024sentencesurvey}. Empirical analyses suggest such auxiliary losses can be brittle and even degrade downstream generalization \citep{liu2019roberta,Aroca-Ouellette2020OnLosses}. In contrast, TG employs no separate encoder or auxiliary sentence-level loss: token and sentence representations are generated using the same set of model parameters and are supervised solely via their contribution to the next-token prediction objective. 
Furthermore, unlike standard transformers, TG enables early layers to attend to a memory of fully contextualized sentence gestalts, directly targeting the contextualization errors identified by \citet{lepori2024racingthoughts}.

\begin{figure*}[t]
  \centering
      \includegraphics[width=0.93\linewidth]{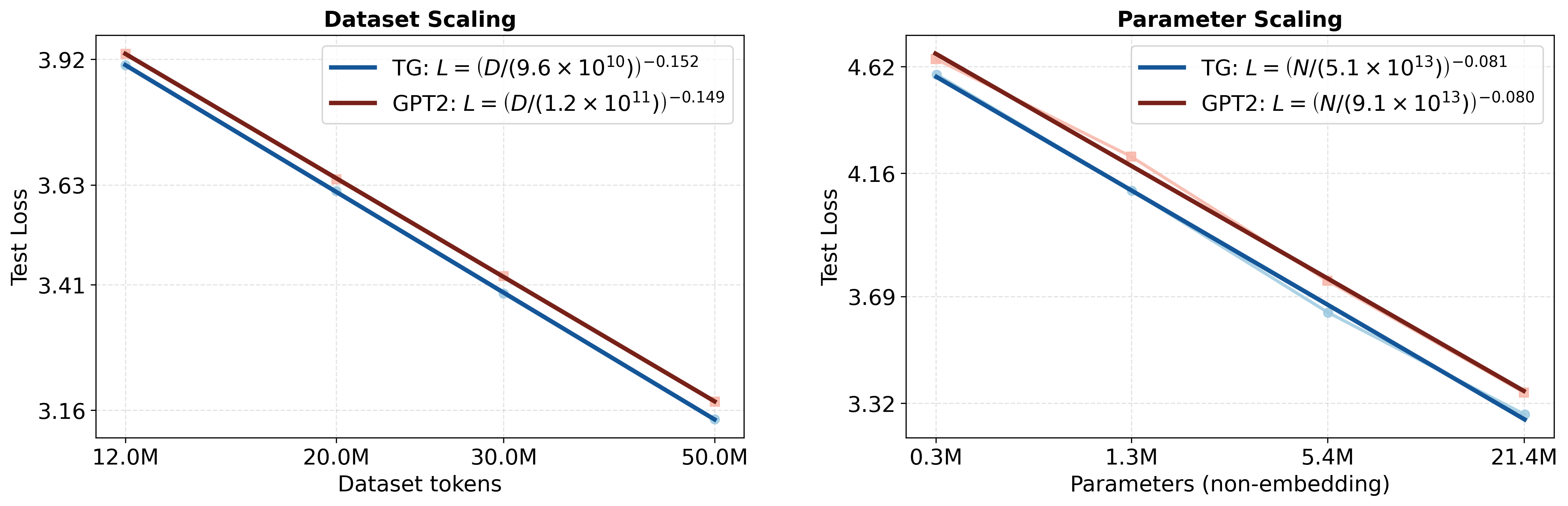}
      \vspace{-0.21cm}

  \caption{\small \textbf{Kaplan-style scaling behavior.} Panel (a) varies training data size $D$ at fixed model size of 85M non-embedding parameters for TG and GPT-2; panel (b) varies model size $N$ at fixed training dataset size of 50M tokens.}
  \label{fig:scaling_combined}
  \vspace{-0.38cm}

\end{figure*}

Empirically, TG yields systematic gains in both learning efficiency and representational robustness. 
TG is consistently more data and parameter efficient than matched GPT-2 baselines pretrained on WikiText-103 (Figure~\ref{fig:scaling_combined}). 
In dataset-scaling experiments ($12$--$50$M training tokens; $N \approx 85$M non-embedding parameters), TG achieves $2$--$4\%$ lower test perplexity at every scale (e.g., $23.2$ vs.\ $24.0$ at $50$M), corresponding to an effective $5$--$8\%$ reduction in the training tokens needed to reach the same loss. 
In parameter scaling experiments at fixed $D=50$M tokens, TG preserves an advantage across a $0.3$M--$21.3$M model size range, where GPT-2 would require $1.33$--$1.42\times$ more parameters to match TG. 
Comparisons to alternative baselines-including adding sentence-boundary bias to GPT-2 via special tokens, replacing TG's sentence-based recurrence with fixed token spans, and sentence-level gisting-style attention masking-show that TG's gains depend on conditioning token generation on semantically coherent, fully contextualized units; performance degrades when sentences are replaced with arbitrary token blocks or when  access to prior tokens is restricted through compression tokens without recurrence (Figure~\ref{fig:baseline_panels}).
Additionally, in an in-context father--son probe, TG improves relational-direction generalization substantially faster than GPT-2, as measured by the likelihood of the target token when the completion query reverses the direction of the prompt (Figure~\ref{fig:reversal_fatherson}).
Finally, ablations confirm that the working memory of sentence gestalts and retaining gradient flow through memory are essential for TG’s performance and that adding per-layer self and cross attention capacity further improves performance (Table~\ref{tab:ablation_summary}).

\section{Main Design Principles}

\begin{figure}[t]
\centering
\includegraphics[width=0.85\linewidth]{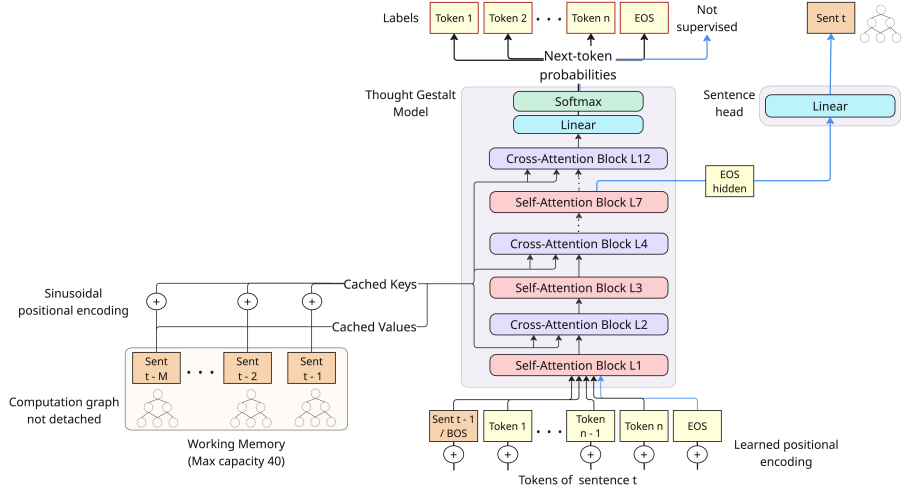}
\caption{\textbf{TG architecture.} Alternating self-attention and cross-attention blocks compose the main model stack. The token view illustrates a non-final sentence in a document (no \textit{EOD} token), and padding tokens are omitted for clarity. Contextualized hidden state of EOS at layer~7 is projected by a small linear sentence head to produce the sentence representation (blue path).}
\label{fig:tg-arch}
\end{figure}

The Thought Gestalt (TG) model learns language as a sequence of sentence-level thoughts using a single transformer stack that functions both as a token decoder and a sentence encoder (Figure~\ref{fig:tg-arch}). Concretely, TG processes a document as a sequence of sentence steps. 
At sentence step  $t\in\{1,\dots,T\}$, where $T$ is the number of sentences in the sentence stream, TG (i) predicts the tokens of sentence $t$ by interleaving causal self-attention over its tokens and cross-attention to a working memory of earlier contextualized sentence vectors; and (ii) compresses sentence $t$ into a single vector, $\mathbf{s}_t$, which is written to memory without detaching gradients. This section describes the TG architecture, its read/write interface to the working memory, and the training and data preparation pipelines that enable stable, scalable learning. For details on batch construction and stochastic regularization, refer to Appendix~\ref{app:add_desings}; for ablations of key TG components see \S\ref{sec:results-ablations}, and for training-time analyses of memory use and gradient flow see Appendix~\ref{app:memory_gate_plots} and Appendix~\ref{sec:comp_grad}.

\subsection{Data Preparation}
\label{sec:data-pipeline}

Our data pipeline converts raw corpora into sentence-level training samples aligned with TG’s computation model. Specifically, we: (i) split datasets into document-level text units based on boundaries provided by the source (i.e., a single Wikipedia article); (ii) split documents into sentences; (iii) enclose each sentence with special boundary tokens; (iv) slice documents into continuous sentence streams of fixed maximum length to create training examples with bounded gradient-graph depth; and (v) construct batches by sampling uniformly from sentence streams, constrained by a target number of lexical tokens per batch, to stabilize GPU memory usage and optimization (detailed in Appendix ~\ref{app:batch-construct}). 

\paragraph{Sentence splitting.}
We preprocess the training corpora by first separating each document (a standalone Wikipedia article) based on title formatting, and then splitting text into sentences using the “SaT Capped'' method. SaT Capped predicts sentence boundaries at the token level and is robust to imperfect punctuation; it enforces a maximum sentence length by applying punctuation-aware fallback rules to split sentences longer than a given maximum token length (set to $L\!=\!64$ tokens in our experiments) into shorter, semantically coherent spans~\citep{Frohmann2024SaT,Minixhofer2023WtP,LCM2024}. We selected this method as it achieves the highest AutoBLEU scores for sentence-level reconstruction, as reported in the Large Concept Model study~\citep{LCM2024}.

\paragraph{Sentence tensorization and boundary markers.}
After sentence splitting, each sentence is tokenized (tokenizer/vocabulary shared across all baselines) and padded to the set maximum sentence length $L_{\max} = 64$ tokens. We then form sentence tensors by adding explicit boundary markers:
a start-of-sentence token \texttt{<BOS>} and an end-of-sentence token \texttt{<EOS>}. The \texttt{<BOS>} token enables the TG model to predict the first lexical token of a new sentence; it sets the context to a non-empty state to enable cross-attending to the working memory. The \texttt{<EOS>} token marks the end of sentence generation and serves as the designated position for extracting the hidden state used to form the sentence representation, $\mathbf{s}_t$ (see \S\ref{sec:model}).
In addition, we reserve a dedicated slot immediately before \texttt{<EOS>} for an end-of-document marker \texttt{<EOD>}. This slot is only added for the final sentence of a document to signal the end of text generation and is \texttt{<PAD>} for all non-final sentences.
Thus, every sentence tensor has the same fixed total length $
L \;=\; 1 + L_{\max} + 2 \;=\; 67.$

\paragraph{Sentence streams as training examples and batch construction.}
TG retains the computation graph of sentence representations written to memory to learn sentence encodings via the next-token prediction loss. At sentence step $t$, the model cross-attends only to the most recent $M$ sentence vectors in the forward pass. However, because the computation graph of memory entries is retained, the computation graph of stored sentence vectors retain links to the graph of earlier sentences they cross-attended to when they were formed. As a result, gradients from step $t$ can recursively backpropagate into the computation graph of older sentence representations that are no longer present in the working memory (Appendix Fig.~\ref{fig:tg_grad}). Therefore, to keep training tractable and avoid storing computation graphs of unbounded depth, we slice training documents (not validation or test documents) into contiguous \emph{sentence streams} with a maximum length cap of $S$ sentences. Each stream functions as an independent training example, and the memory is reset between streams, which truncates the gradient chain. This strategy preserves long-range conditioning while capping the maximum depth of backpropagation. To maintain optimization stability given the variable number of lexical tokens per stream, batches are constructed by uniformly sampling streams until a target range of lexical tokens per batch is reached (details in Appendix~\ref{app:batch-construct}).

\subsection{Model}\label{sec:model}

\paragraph{Sentence representation.}
Let $x_{1:L}$ be the tokens of the current sentence, $\mathbf{H}^{(\ell)} \in \mathbb{R}^{L\times d}$ denote the hidden states at layer $\ell$ of the model stack
(with $d=d_{\text{model}}$), and  $i_{\text{EOS}}$ index the \texttt{<EOS>} position.
At a fixed mid layer $\ell_s$ ($\ell_s=7$ in our setting), we form the sentence vector
\vspace{0.2cm}
\[
\mathbf{s}_t \;=\; W_{\text{sent}}\, \mathbf{H}^{(\ell_s)}_{i_{\text{EOS}}} \in \mathbb{R}^{d}
\vspace{0.2cm}
\]
Here $W_{\text{sent}}$ denotes the sentence head, a single linear layer
$W_{\text{sent}}\!\in\!\mathbb{R}^{d\times d}$ (depth~1). At the end of step~$t$, the computed $\mathbf{s}_t$ is appended to the working memory, without detaching its compute graph, and if working memory is full, the oldest entry is removed.
Importantly, $\mathbf{s}_t$ lives in the same $d$-dimensional space as token hidden states
($d=768$ in our experiments).
Choosing a mid layer to extract the hidden state used for generating the sentence representation, $\mathbf{s}_{t}$, is motivated by prior evidence showing that intermediate transformer layers carry the most contextual and transferable features, whereas top layers increasingly specialize in token-level decisions (i.e., next-token lexicalization) \citep{geva2021ffnkv, Liu2023LostInTheMiddle, rogers2020primer}.

\paragraph{Cross-attention to working memory.}
TG maintains a rolling working memory of fixed capacity $M$. In our experiments, we set $M=40$ to approximate the 1024-token context window of the GPT-2 baseline (the average sentence length in our training corpus is $\sim$25 tokens).
At step $t$, the working memory contains the most recent sentence representations
$\mathbf{s}_{t-M_t}, \ldots, \mathbf{s}_{t-2}, \mathbf{s}_{t-1}$, where
$M_t = \min(M, t-1)$, ordered from oldest to most recent. For efficiency, we form (once per sentence step) the unprojected memory key/value matrices, which are reused by all cross-attention layers when processing sentence $t$ 
\vspace{0.2cm}
\[
K_M \;=\; \big[\mathbf{s}_{t-M_t},\ldots,\mathbf{s}_{t-1}\big] \;+\; P^{(\text{sent})}_{1:M_t}
\;\in\; \mathbb{R}^{M_t\times d},
\qquad
V_M \;=\; \big[\mathbf{s}_{t-M_t},\ldots,\mathbf{s}_{t-1}\big]
\;\in\; \mathbb{R}^{M_t\times d},
\vspace{0.2cm}
\]
where $P^{(\text{sent})}_{1:M_t}$ are sinusoidal sentence-index positional encodings added to keys only.
Each cross-attention block has its own learned projections, thus at cross-attention layer $\ell$ we use
\vspace{0.2cm}
\[
Q^{(\ell)}=\mathbf{H}^{(\ell)} W_Q^{(\ell)},\quad
K_M^{(\ell)}=K_M W_K^{(\ell)},\quad
V_M^{(\ell)}=V_M W_V^{(\ell)}.
\vspace{0.2cm}
\]
Multi-head cross-attention is then computed per head (with $n_{\text{head}}{=}12$ and
$d_h{=}d/n_{\text{head}}$) and concatenated in the standard way; in compact form,
\vspace{0.2cm}
\[
\mathrm{Attn}\!\big(Q^{(\ell)},K_M^{(\ell)},V_M^{(\ell)}\big)
=
\mathrm{softmax}\!\left(\frac{Q^{(\ell)} \big(K_M^{(\ell)}\big)^{\!\top}}{\sqrt{d_h}}\right)
V_M^{(\ell)}.
\vspace{0.2cm}
\]

Injecting sentence order information into keys (rather than values) decouples positional information from semantic content and encodes position directly into attention scores, which prior work shows can improve
robustness under length variation~\citep{press2021shortformer,press2022alibi,miyazaki-etal-2024-understanding}.
In TG, token queries already include token-level positional information via $\mathbf{P}^{(\text{tok})}$, thus
we add positional structure only to the memory keys.
Appendix~\ref{sec:comp_grad} provides evidence that, as pretraining scale increases, a
larger fraction of the training signal flows through the cross-attention pathway, consistent with the analysis 
Appendix~\ref{sec:comp_grad} showing that the
model learns to rely increasingly on the sentence-memory as training progresses and sentence representations stabilize.

\paragraph{Token embeddings and self-attention.}
We use a standard token embedding matrix $\mathbf{E}\!\in\!\mathbb{R}^{|\mathcal{V}|\times d}$ and learned token positional embeddings $\mathbf{P}^{(\text{tok})}\!\in\!\mathbb{R}^{L\times d}$; the input to the model stack is $\mathbf{X}^{(0)}=\mathbf{E}[x_{1:L}]+\mathbf{P}^{(\text{tok})}$.
Each self-attention layer is pre-norm, causal multi-head attention with residual connections.
Within a block~$\ell$, token self-attention uses its own projections $W_Q^{(\ell)},W_K^{(\ell)},W_V^{(\ell)}$ and standard scaled dot-product attention.

\paragraph{Learnable memory gates.}
Each cross-attention layer has a scalar, learnable memory gate $g_{\text{mem}}^{(\ell)}$ that scales the cross-attention increment before it is added back via the residual path.
These gates let the model modulate reliance on the memory pathway over training (up-weighting memory once sentence representations stabilize (Appendix~\ref{app:memory_gate_plots})).

\paragraph{Context seeding.}
In standard transformers, the prediction of the first token relies on a static \texttt{<BOS>} embedding, lacking local context. Because TG resets the context window at every sentence boundary, this issue is exacerbated.
To address this, we replace the static \texttt{<BOS>} embedding for step $t$ with the preceding sentence representation $\mathbf{s}_{t-1}$, if available:
\vspace{0.2cm}
\[
\mathbf{X}^{(0)}_{i_{\text{BOS}}} \leftarrow \mathbf{s}_{t-1}.
\vspace{0.2cm}
\]
This transforms the start token into a contextualized state. Consequently, the first lexical token of a new sentence can access prior context via two complementary pathways: (1) cross-attention to the memory, and (2) local self-attention to the \texttt{<BOS>} position, which now explicitly encodes the previous sentence representation.

\subsection{Training schedules}
\label{sec:training-schedules}

\paragraph{Sentence-stream curriculum}
Retaining computation graphs through memory writes causes the effective backpropagation depth to grow with the number of sentences processed continuously (i.e., without resetting the memory between sentence streams), even though the number of sentence representations available in forward pass is capped at $M=40$. Early in training, sentence representations are largely uninformative; long sentence streams therefore increase compute and optimization difficulty without providing useful long-range credit assignment. We address this with a curriculum over the maximum sentence-stream length $S$ used to slice training documents: we begin with shorter streams and periodically increase $S$ and re-chunk the training split (validation and test sets are never split into sentence streams). In our experiments with memory capacity $M{=}40$, we start at $S{=}30$ sentences and increase by $+12$ every 5 epochs. This expands the effective dependency span of optimization from $\sim$750 (average sentence length of $\sim25{\times}30\!\approx\!750$)  to $\sim$2000 tokens over the course of training, while keeping early optimization focused on within-sentence token modeling and short-range memory use.

\paragraph{Down-weighting frequent boundary token (EOS).}
In TG, end-of-sentence markers (\texttt{<EOS>}) are more frequent than lexical tokens, as they appear at the end of every sentence and are often comparatively easy to predict (strongly signaled by punctuation and syntax). To mitigate this frequency and easy-label imbalance, we apply a frequency-aware reweighting of the token-level loss that down-weights \texttt{<EOS>} targets after an initial warm-up epoch (weight 1.0 for epoch~1, then 0.05 thereafter) \citep{lin2017focal,cui2019classbalanced}. This affects the training loss the model receives; separately, special-token targets (\texttt{<EOS>}/\texttt{<EOD>}) are excluded from reported perplexities and checkpoint selection, as described in \S\ref{sec:setup}.

\section{Results}
\label{sec:results}

We evaluate the Thought Gestalt model through a series of experiments designed to measure data efficiency and representational robustness. First, we analyze TG's \emph{scaling behavior} by measuring how loss improves with training data and model size relative to a standard GPT-2 baseline, following the empirical scaling-law framework of \citet{kaplan2020scaling}. Second, we benchmark TG's data scaling against three architectural baselines that isolate key design concepts: (1) \emph{GPT-2 with Sentence Boundary Bias}, which induces structural bias by retaining sentence start/end tokens (\texttt{<BOS>}, \texttt{<EOS>}) in the token stream; (2) \emph{Fixed Token-Span Recurrence}, which replaces TG's sentence-level processing with fixed-length token windows similar to prior recurrent work \citep{bulatov2022rmt, hutchins2022blockrecurrent}; and (3) \emph{GPT-2 + Gist Masking}, an in-context gisting-style attention masking method \citep{mu2023gisting}, where tokens attend causally within a sentence and access preceding sentences only through compressed ``gist'' tokens that are computed in a single pass, without recurrence. Third, we assess \emph{relational direction generalization}, which evaluates robustness to the in-context reversal curse given a controlled father--son probe \citep{berglund2024reversal,lin2024delving}. Finally, we conduct \emph{design ablations} to identify how specific components, such as gradient flow through memory, drive performance gains.

\subsection{Experimental Setup}
\label{sec:setup}
All models are pre-trained on fixed subsets of the WikiText-103 training split, holding the validation and test sets fixed across experiments \citep{merity2016pointer}. All reported losses and perplexities are computed only over \emph{lexical} tokens: we exclude special-token label positions (e.g., the position predicting frequent end-of-sentence and end-of-document tokens, \texttt{<EOS>} and \texttt{<EOD>} respectively) from both reporting and checkpoint selection. We optimize with AdamW ($\beta_1{=}0.9$, $\beta_2{=}0.999$, weight decay $0.01$) using a peak learning rate of $2.5\times10^{-4}$ and a cosine schedule with 2\% warmup. The best model checkpoints with lowest validation perplexity are selected and reported. During training, we early-stop based on validation perplexity with minimum  $\Delta$ of $=0.1$ perplexity and patience of $3$ epochs.

\subsection{Scaling Efficiency} 
\label{sec:results-scaling}

We quantify the data and parameter efficiency of Thought Gestalt (TG) using the empirical scaling-law framework of \citet{kaplan2020scaling}. Following this work, we measure $L$ the held-out test cross-entropy loss in nats/token and track model size as $N$, the number of \emph{non-embedding} parameters. When other factors are not the limiting bottleneck, Kaplan et al.\ observed approximate power-law relationships
$L(N)\approx (N_c/N)^{\alpha_N}$ and $L(D)\approx (D_c/D)^{\alpha_D}$, which appear as straight lines in log--log space. In this regime, architectural improvements typically manifest as a downward shift of the curve (lower loss at fixed resources). We first fit these laws to TG and a matched GPT-2 baseline trained under identical optimization and evaluation protocols described in \S\ref{sec:setup} to compare their data and parameter efficiency.

\paragraph{Training dataset scaling.}
We measure scaling with dataset size by training TG and GPT-2 on WikiText-103 subsets ranging from $12$M to $50$M tokens while holding model capacity fixed at $N\approx85$M non-embedding parameters (12 layers, $d_{\text{model}}{=}768$) for both models.
Using a relatively large fixed-$N$ model helps isolate how performance changes with $D$ before capacity limitations dominate, analogous to the dataset-scaling analyses in \citet{kaplan2020scaling}.

Figure~\ref{fig:scaling_combined}(a) shows that TG achieves lower test perplexity at every dataset size. Fitting a power law to the test loss yields similar scaling exponents for the two models (TG: $\alpha\approx0.152$; GPT-2: $\alpha\approx0.149$), suggesting that over this range the gain is primarily due to an intercept shift. Across our tested range this corresponds to a $2$--$4\%$ reduction in perplexity (e.g., at $D{=}50$M tokens, TG reaches 23.2 test perplexity versus 24.0 for GPT-2; see Appendix~\ref{app:scaling_tables} for other perplexities).
To express the improvement as data efficiency, we compute an effective data multiplier $m_D = D_{\text{GPT-2}}(L_{\text{TG}}(D))/D$,
where $L_{\text{TG}}(D)$ is TG's achieved test loss at dataset size $D$, and $D_{\text{GPT-2}}(L)$ is the number of training tokens GPT-2 would need to reach loss $L$ according to the fitted GPT-2 scaling curve. We find $m_D\approx1.05$--$1.08$, meaning GPT-2 requires about 5--8\% more training tokens to match TG's loss; supporting the hypothesis that learning latent representations at more abstract levels than tokens can lead to more sample efficient learning.

\paragraph{Model size scaling.}
To assess parameter efficiency, we fix the training set at $D{=}50$M tokens and vary model size from $N\approx0.34$M to $21.3$M non-embedding parameters.
We hold depth fixed at 12 layers and scale width ($d_{\text{model}}\in\{48,96,192,384\}$), which provides a controlled logarithmic sweep over $N$. This choice follows \citet{kaplan2020scaling}, who observed that transformer loss depends primarily on total non-embedding parameter count and only weakly on the model depth or width within a broad, non-extreme range. Figure~\ref{fig:scaling_combined}(b) shows that TG attains lower perplexity than GPT-2 at every model size. Power-law fits to test loss yield nearly identical exponents (TG $\alpha\approx0.081$; GPT-2 $\alpha\approx0.080$), again consistent with an intercept shift.
Interpreted as an effective parameter multiplier, 
$m_N = N_{\text{GPT-2}}(L_{\text{TG}}(N))/N$, the fitted curves imply that matching TG’s loss would require roughly $1.33$--$1.42\times$ more GPT-2 parameters over the tested range. For example, under the fit, matching TG at $N\approx21$M would require $\approx30$M GPT-2 non-embedding parameters.

\begin{figure*}[t]
  \centering
  \begin{minipage}[t]{0.33\linewidth}
    \centering
    \includegraphics[width=\linewidth]{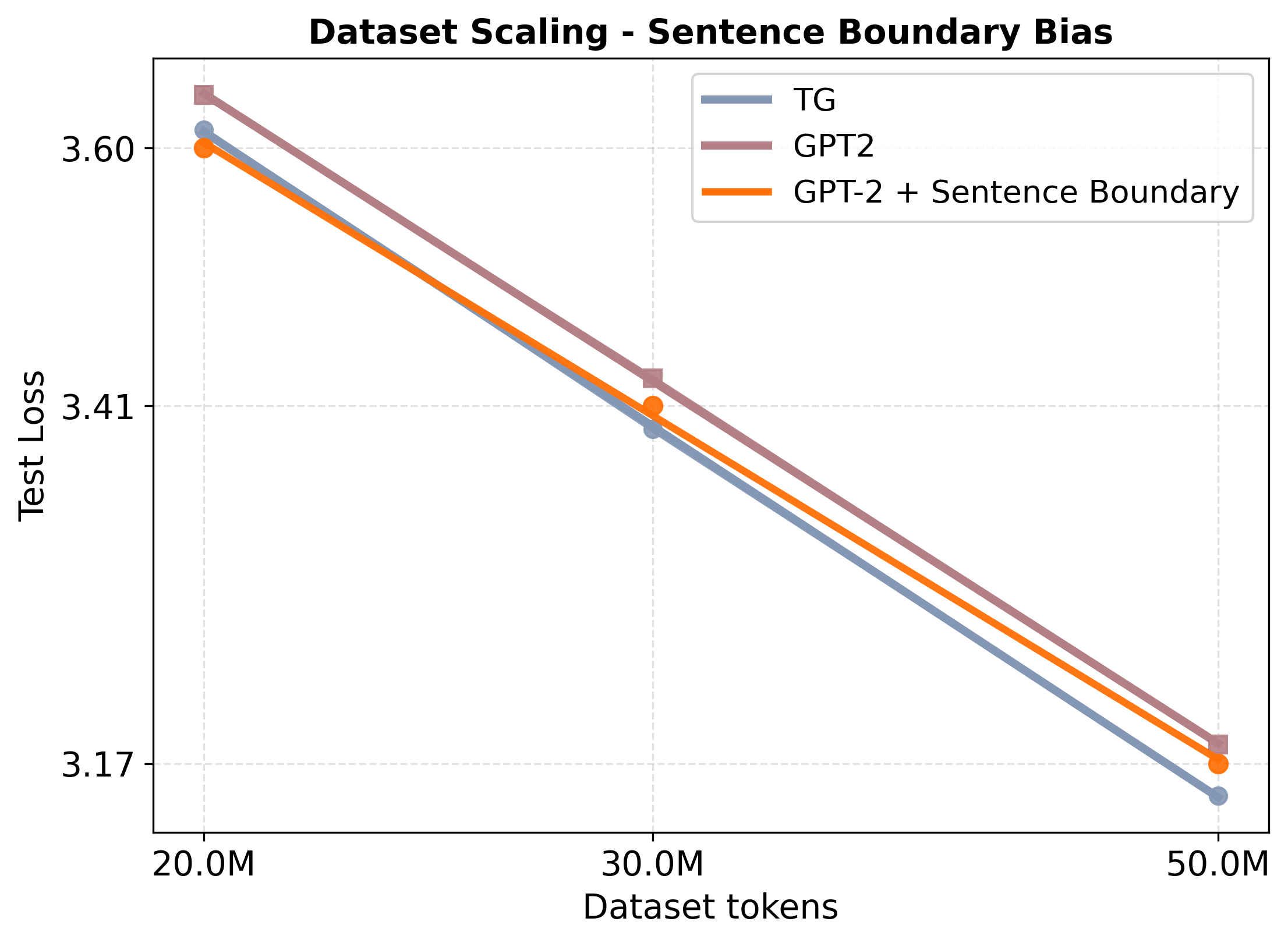}
    
  \end{minipage}\hfill
  \begin{minipage}[t]{0.33\linewidth}
    \centering
      \includegraphics[width=\linewidth]{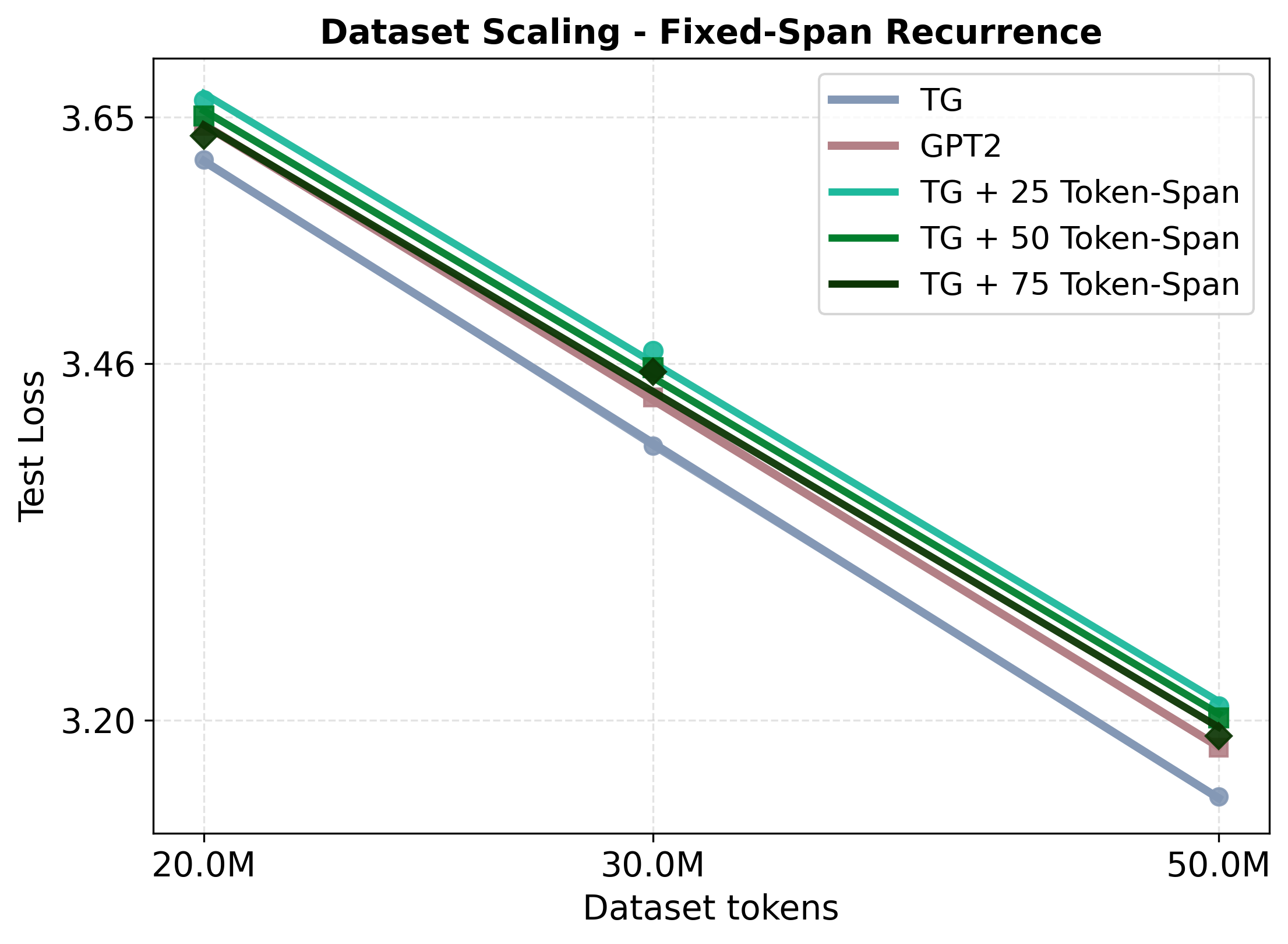}
    
  \end{minipage}\hfill
  \begin{minipage}[t]{0.33\linewidth}
    \centering
      \includegraphics[width=\linewidth]{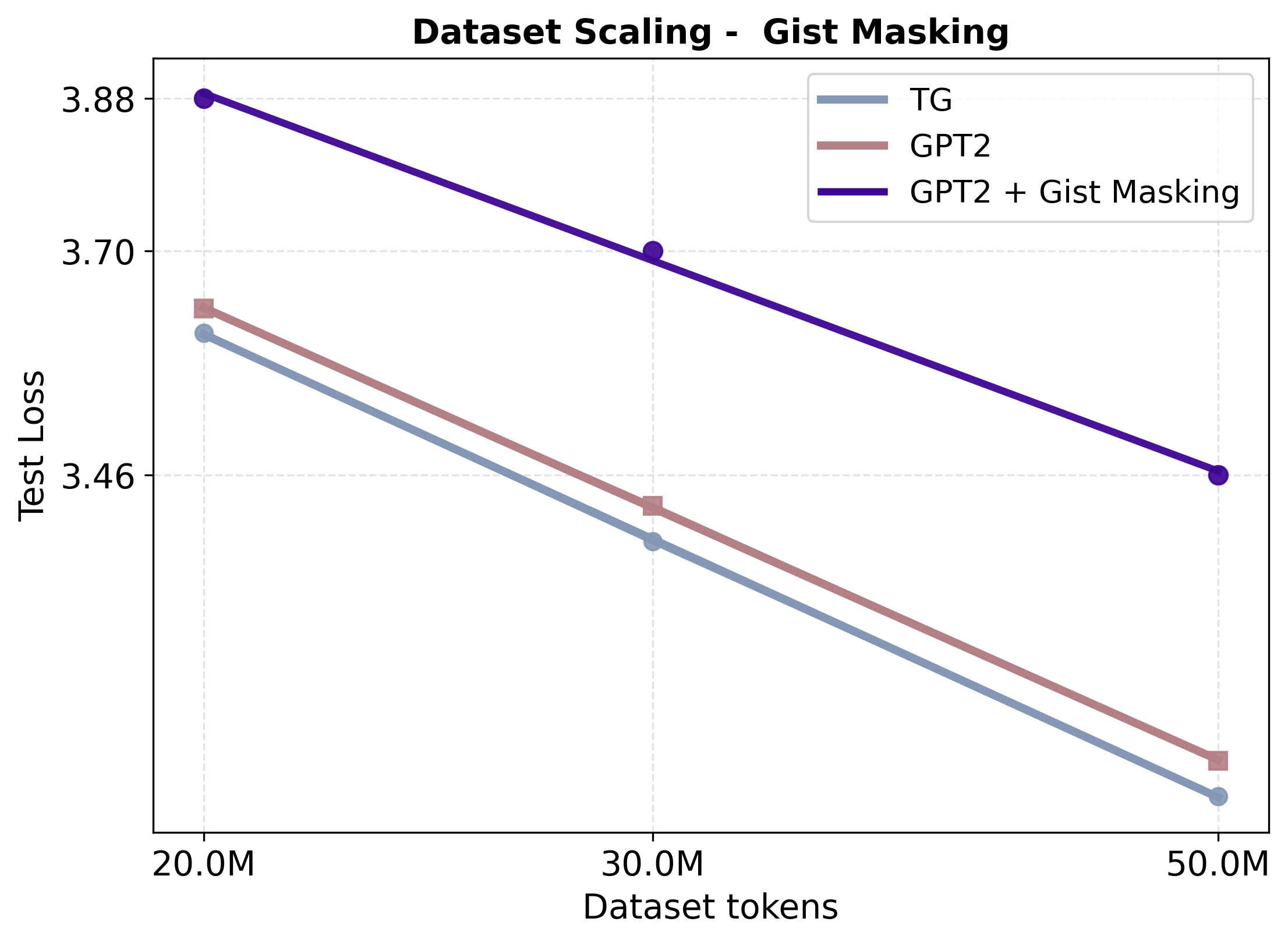}
    
  \end{minipage}
  \caption{\small \textbf{Other baseline data-scaling comparisons.} All panels report test cross-entropy loss (nats/token; lexical tokens only) vs.\ training tokens at $D\in\{20\mathrm{M},30\mathrm{M},50\mathrm{M}\}$ under identical splits and early stopping. Panel (a) isolates the effect of explicit sentence boundary markers in the token stream of a standard decoder model (GPT-2); panel (b) replaces each sentence input with fixed size token-span in TG, disregarding sentence boundaries; panel (c) compares to an in-context gist-masking baseline where sentence representations are computed in a single pass without recurrence \citep{mu2023gisting}.}
  \label{fig:baseline_panels}
\end{figure*}

\subsection{Other Baselines}

\paragraph{GPT-2 with sentence boundary bias.}
TG leverages sentence boundaries as a structural bias to process information in semantically coherent units. Here, we evaluate whether a standard transformer benefits simply from making this structure explicit in the token stream. To do this, we take the sentence-segmented output of the TG data pipeline (\S\ref{sec:data-pipeline}) and flatten it into a continuous sequence for a standard GPT-2 model. Crucially, we retain the \texttt{<BOS>} and \texttt{<EOS>} markers to explicitly delimit boundaries, while removing the padding tokens required for TG’s fixed-size sentence tensors. In this setup, \texttt{<BOS>} predicts the sentence’s first lexical token, the last lexical token predicts \texttt{<EOS>}, and \texttt{<EOS>} predicts the \texttt{<BOS>} of the subsequent sentence. The model architecture and attention mechanism remain identical to the GPT-2 baseline. To avoid making the comparison hinge on predicting frequent boundary events, we compute validation loss for checkpoint selection and report perplexity over \emph{lexical} tokens only, excluding special-token label positions, matching the TG evaluation protocol (\S\ref{sec:setup}). Figure~\ref{fig:baseline_panels}(left) shows that inducing sentence boundary bias improves GPT-2 performance across all dataset sizes, with the most significant gains in the low-data regime (test PPL decreases $38.1\!\rightarrow\!36.6$ at 20M; see Table~\ref{tab:baseline_sentence_boundary}). However, the improvement shrinks with scale (e.g., $24.0\!\rightarrow\!23.7$ at 50M), and TG surpasses this baseline once more data is available (TG: $23.2$ vs.\ $23.7$ at 50M). This indicates that boundary tokens alone provide a helpful structural inductive bias for token LMs, but this is not sufficient to replicate the efficiency gains derived from TG's learning and reuse of contextualized sentence-level latent states.

\paragraph{Fixed token-span recurrence.}
Many recurrent and memory-based transformers propagate information by processing fixed-length token blocks and caching hidden states without aligning blocks to linguistically meaningful units (e.g., transformer-XL, block-recurrent transformers, and recurrent memory-token schemes; \citep{dai2019transformerxl,hutchins2022blockrecurrent,wu2022memorizing}). 
To determine if TG's performance gains are attributable specifically to using semantically coherent text segments for recurrence/compression, here we evaluate TG with an identical architecture and training procedure, but replace sentence segmentation with fixed token spans of length $N\in\{25,50,75\}$ lexical tokens (with $N{=}25$ chosen to match the average sentence length in our corpus). Each span is treated as a TG “sentence step”: it is wrapped with boundary tokens and compressed into one memory vector. The only difference is that memory entries now summarize arbitrary token windows rather than syntactically/semantically coherent sentences. As shown in Figure~\ref{fig:baseline_panels} (middle) and Appendix Table~\ref{tab:baseline_token_span}, fixed-span recurrence underperforms sentence-based TG across all data sizes; for example, at $50$M training tokens TG reaches $23.2$ test PPL, while the $25$/$50$/$75$-span variants reach $24.7/24.5/24.2$, respectively. Increasing the span length helps, but does not close the gap to TG: even the best model ($75$ tokens) remains worse than TG at every scale and performs more comparably to standard GPT-2. This suggests that TG’s advantage is driven by compressing semantically coherent units: sentence boundaries provide superior targets for event segmentation and stable memory organization compared to arbitrary blocks, a finding consistent with cognitive theories of comprehension and memory \citep{radvansky2011event,jarvella1971syntactic,zwaan1998situation}.

\paragraph{GPT-2 + gist masking.}
TG’s attention pattern—full causal token interaction within a sentence, and access to earlier content only through compact sentence vectors—is similar to gisting: tokens attend to their current segment and can only access prior segments through a small set of “gist” representations learned via attention masking in a standard transformer \citep{mu2023gisting}. To test whether this attention distribution mechanism can reproduce TG’s gains without recurrence or an external memory, we implement a GPT-2 baseline that (i) inserts \texttt{<BOS>}/\texttt{<EOS>} boundaries in the token stream and (ii) applies an additive attention-bias mask that restricts each token to attend causally within its own sentence, while accessing previous sentences only via each sentence’s last token, \texttt{<EOS>}, acting as a “gist” token. This matches TG’s attention connectivity, but removes recurrence, which has a key stabilizing effect: TG exposes fully contextualized sentence vectors to every layer via cross-attention, whereas GPT-2 + gist masking forces layers to read from in-context compression tokens whose representations are formed in parallel with all other tokens. In the original work, a prompt prefix is compressed into learned gist tokens which are computed in a separate pass for caching (a one iteration recurrence); here we compute all sentence gists in a single pass, without caching, to ablate recurrence. Empirically, the gist-masking baseline performs worse across all data scales (test PPL 31.9 at 50M; Table~\ref{tab:baseline_gisting}), far behind both GPT-2 (24.0 at 50M) and TG (23.2 at 50M), as seen in Figure~\ref{fig:baseline_panels} (right). These results show that in-context compression, restricting direct access to prior token, can degrade pretraining when compressed representations are formed concurrently with the later tokens that depend on them, without recurrence.

\begin{figure*}[t]
  \centering

  \begin{minipage}{0.6\linewidth} 
    \centering
    \includegraphics[width=\linewidth]{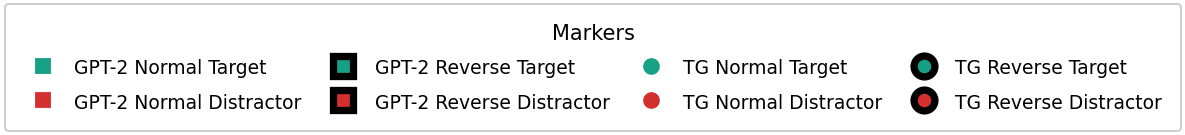}
  \end{minipage}
  \vspace{0.5em}

    \begin{minipage}[c]{0.4\linewidth} 
    \centering
    \includegraphics[width=\linewidth]{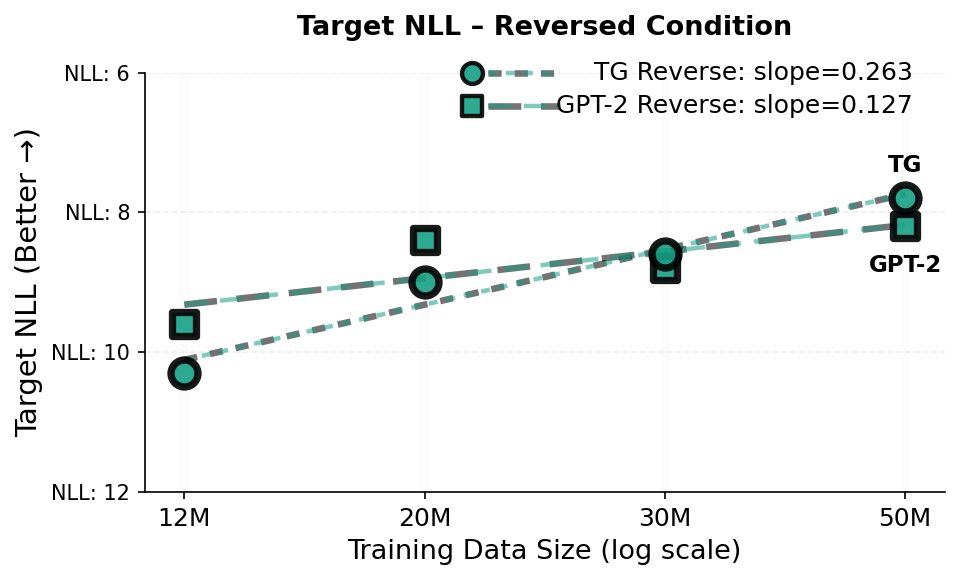}
  \end{minipage}
  \hspace{0.08\linewidth} 
  \begin{minipage}[c]{0.4\linewidth} 
    \centering
    \includegraphics[width=\linewidth]{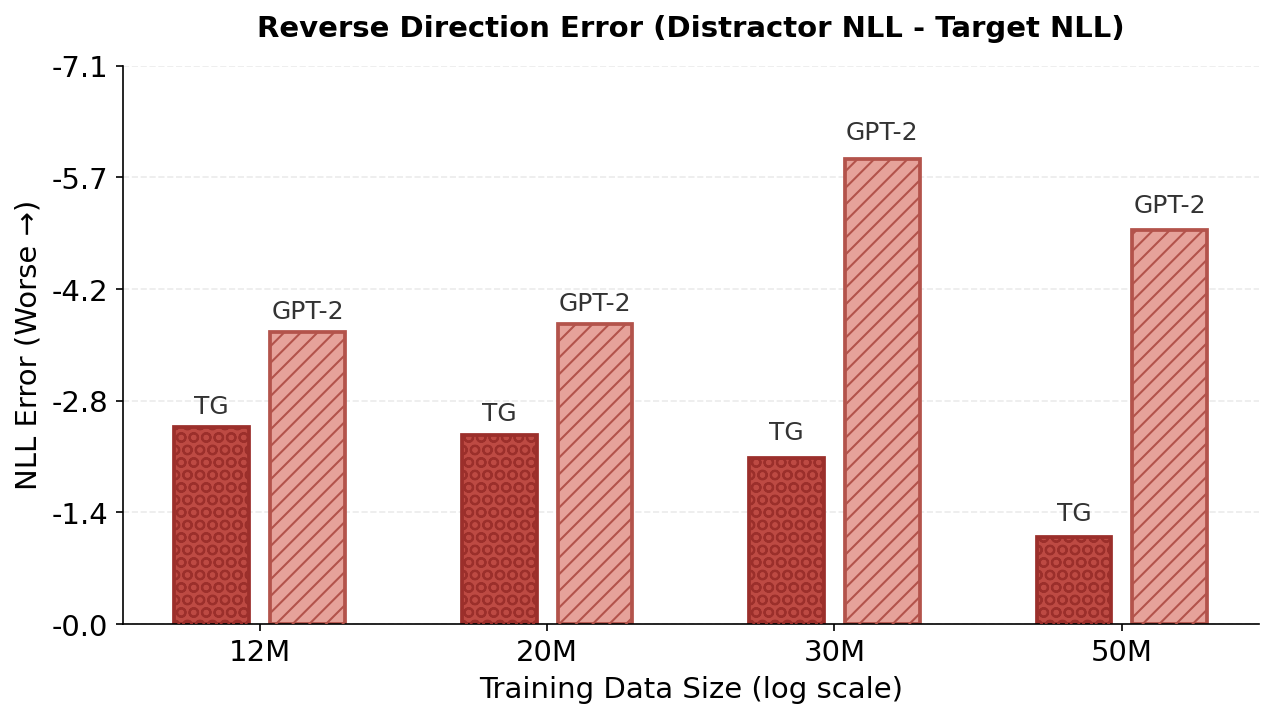}
  \end{minipage}
  
  \vspace{0.8em}

  \begin{minipage}[c]{0.53\linewidth}
    \centering
    \includegraphics[width=\linewidth]{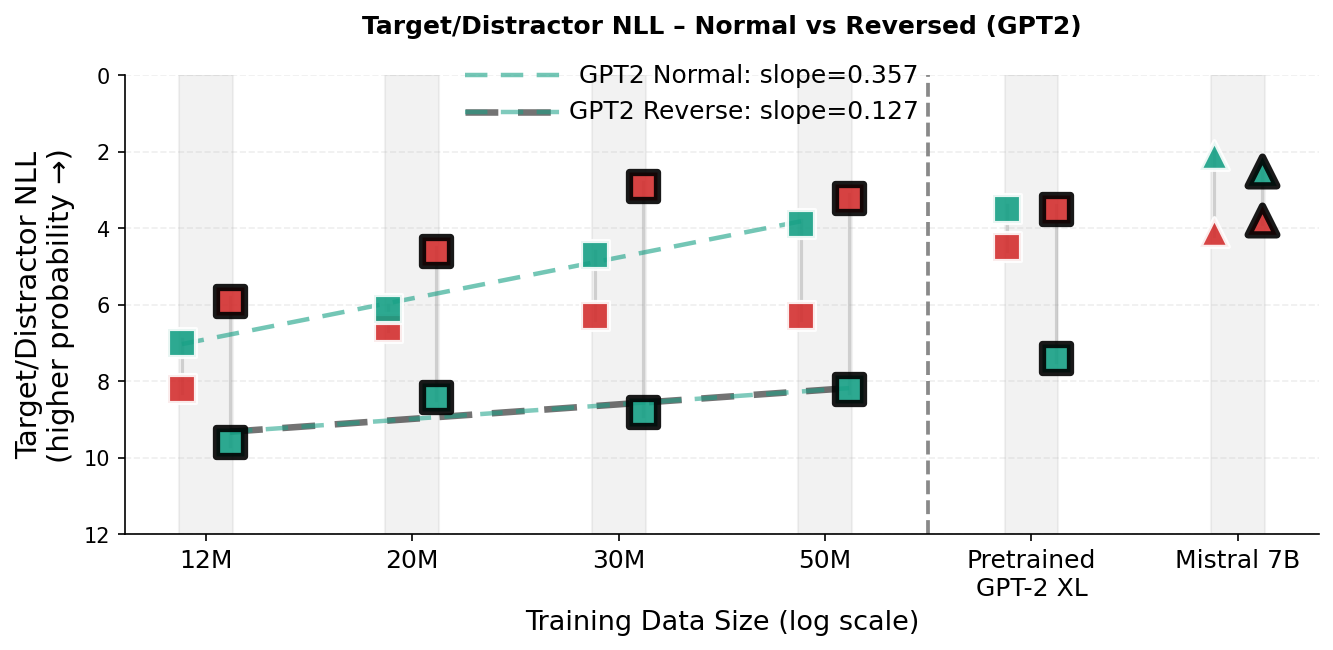}
  \end{minipage}
  \hspace{0.02\linewidth} 
  \begin{minipage}[c]{0.4\linewidth}
    \centering
    \includegraphics[width=\linewidth]{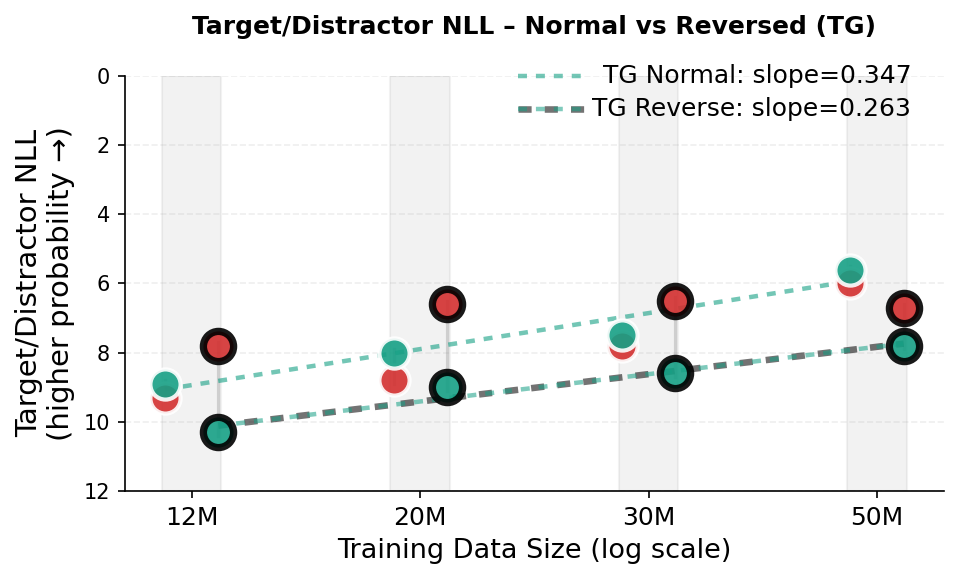}
  \end{minipage}


  \caption{\textbf{Father--son in-context reversal curse probe.}
Each prompt states a relation (e.g., ``The son of Michael is John.'') and then queries either the same direction (Normal: ``The son of Michael is'') or the inverse (Reversed: ``The father of John is'').
We report mean NLL (nats) at the first answer position for the correct \emph{target} name (green) and the \emph{distractor} name that also appears in the prompt (red).
Top-left: target-token NLL in the Reversed condition vs. training dataset size reported to compare relational direction generalization of TG and GPT-2 and the rate of improvement as training data increases (dashed lines are least-squares fits).
Top-right: reverse-direction margin $\Delta=\mathrm{NLL}(\text{distractor})-\mathrm{NLL}(\text{target})$, where $\Delta<0$ indicates an in-context reversal error (distractor has higher probability).
Bottom-left: target/distractor breakdown for GPT-2, including the pretrained Hugging Face GPT-2 XL checkpoint (1.5 billion parameters; pretrained on $\sim$9B WebText tokens) and Mistral 7B (7 billion parameters) to show performance of scaled models.
Bottom-right: the same breakdown for TG.}
  \label{fig:reversal_fatherson}
\vspace{-0.5cm}
\end{figure*}

\subsection{Reversal curse evaluation (Father--Son completion).}
\label{sec:results-reversal}

Prior work on the reversal curse characterizes a training-time directional asymmetry: models trained (or fine-tuned) on relational statements in one direction (``A is B'') often fail to generalize to the inverse (``B is A'') \citep{berglund2024reversal,lin2024delving}.
Here we evaluate a distinct, inference-time phenomenon: an \emph{in-context reversal curse}, where the relation is stated in the prompt, and the completion query asks for either the same direction or its inverse.
\citet{berglund2024reversal} note that reversal curse is not observed when relations are expressed in-context for the large models they evaluate (including GPT-3.5/4) \citep{openai2023gpt4}; we find that a substantial in-context directional bias persists at smaller scales, including the pretrained Hugging Face GPT-2 XL checkpoint (1.5B parameters; WebText pretraining on $\sim$40GB of text, roughly 9 billion tokens) \citep{hf_gpt2xl}. Furthermore, we show that this in-context reversal error is resolved only at the scale of Mistral 7B (7 billion parameters, trained on trillions of tokens) \citep{jiang2023mistral7b}.
To our knowledge, prior work has not documented the in-context directional asymmetry of smaller scale models.

Our \emph{in-context reversal curse} experiments use Father--Son relation probes in completion mode.
Each example consists of a sentence establishing a relationship, followed by a query prefix.
In the \emph{Normal} condition, the query repeats the relational statement up to the answer position (e.g., context: ``The son of Michael is John.''; query: ``The son of Michael is'').
In the \emph{Reversed} condition, the query inverts the relation direction (e.g., context: ``The son of Michael is John.''; query: ``The father of John is''). We evaluate the model token-probability distribution at the first answer position and report the negative log-likelihood (NLL; in nats) of two candidates: the target token (correct answer) and the distractor token (other name present in the prompt).
Figure~\ref{fig:reversal_fatherson} visualizes the results.
In the scatter plots (Fig.~\ref{fig:reversal_fatherson}, top-left and bottom row), the y-axis is inverted so that higher marker placement (lower NLL) indicates higher token probability (higher model confidence).
An in-context reversal error corresponds to the distractor having lower NLL than the target (a red marker above the corresponding green marker), equivalently a negative reverse-direction margin $\Delta=\mathrm{NLL}(\text{distractor})-\mathrm{NLL}(\text{target})<0$ (Fig.~\ref{fig:reversal_fatherson}, top-right).
Dashed lines are least-squares fits of the target-NLL trend across the dataset-scaling checkpoints (the same runs as in Fig.~\ref{fig:scaling_combined}a).

\vspace{-0.1cm}

\paragraph{Reversed condition.}
In the reversed condition, TG improves the probability of target token substantially faster than GPT-2 (Fig.~\ref{fig:reversal_fatherson}, top-left; target-NLL trend slope $0.263$ vs.\ $0.127$) and achieves lower target NLL in the reversed condition at 30M and 50M training dataset sizes. To summarize directional bias, we plot the reverse-direction margin $\Delta \;=\; \log p(\text{target}) - \log p(\text{distractor})
\;=\; \mathrm{NLL}(\text{distractor}) - \mathrm{NLL}(\text{target}),
$
where $\Delta>0$ indicates a preference for the correct answer and $\Delta<0$ indicates a distractor preference (a reversal error; Fig.~\ref{fig:reversal_fatherson}, top-right). In the reversed condition, $\Delta$ improves monotonically for TG (approximately $-2.5\rightarrow -1.1$ nats from 12M$\rightarrow$50M), indicating steadily reduced bias toward the distractor.
In contrast, GPT-2’s margin becomes more negative through 30M and only slightly recovers at 50M, with a strong distractor preference persisting even for the pretrained GPT-2 XL checkpoint (Fig.~\ref{fig:reversal_fatherson}, bottom-left).
This pattern suggests that for a standard token-stream transformer, additional data may amplify an order-bound shortcut (copying the wrong entity at a similar position) rather than capturing the underlying relationship.
TG’s sentence-level contextualization reduces this shortcut pressure and moves the margin toward zero, with trends pointing toward a resolution of the in-context reversal curse at a much lower pretraining scale than GPT-2.

Figure~\ref{fig:reversal_fatherson} (bottom-left) shows that the pretrained GPT-2 XL checkpoint succeeds in the Normal (copy-paste) query (target preferred over distractor) but exhibits a large in-context reversal error in the Reversed query (distractor preferred over target).
In contrast, Mistral 7B (a 7 billion parameter foundation model likely pretrained on order of trillions of tokens \citep{deepseekv3}) prefers the target in both directions, indicating that this in-context directional bias is largely absent at 7B scale \citep{jiang2023mistral7b}.

\paragraph{TG vs. GPT-2: Normal condition.}
In the normal (copy-paste) completion, both models improve at nearly the same rate (Fig.~\ref{fig:reversal_fatherson}, bottom panels; slopes $0.357$ for GPT-2 vs.\ $0.347$ for TG).
GPT-2 is more confident in absolute terms (lower target NLL), while TG shows a consistent offset that shrinks with additional training data.
This is expected, as the normal condition represents a trivial in-context token-sequence copying task; GPT-2 has direct access to the exact token sequence, while TG must infer the sequence from a compressed sentence state.
Critically, TG matches GPT-2’s improvement trend in the normal condition while exhibiting a much higher improvement rate in the inverse condition (Fig.~\ref{fig:reversal_fatherson}, top-left and top-right).
This supports our hypothesis that conditioning on fully contextualized latent states, rather than solely on position-tied token sequences, can mitigate the reliance on surface word order that causes the reversal curse.

\begin{table}[t]
\small
  \centering
  \caption{\textbf{Ablation study on $30\mathrm{M}$-token pretraining.}
  We select checkpoints by validation perplexity (lexical tokens only) and report the corresponding \emph{test} perplexity.
  $N$ denotes the non-embedding parameter count.
  Throughput is measured during training (sent./sec) on a single NVIDIA A40 GPU with no model or data parallelism.}
  \label{tab:ablation_summary}

  \renewcommand{\arraystretch}{1.15}

  \begin{tabular}{@{}lccc@{}}
    \toprule
    \makecell[l]{Ablation}
      & \makecell[c]{\textbf{Test}\\\textbf{PPL}}
      & \makecell[c]{\textbf{Throughput}\\(sent./sec)}
      & \makecell[c]{\textbf{Model Size}\\$N$} \\
    \midrule
    TG (Baseline)
      & 29.8
      & 21
      & \textbf{85.6M} \\
    \midrule
    Layer type: Self $\rightarrow$ Cross
      & \textbf{29.4}
      & 17
      & 114M \\
    \midrule
    \makecell[l]{Layer type: parallel Self \& Cross}
      & 29.7
      & 17
      & 114M \\
    \midrule
    \makecell[l]{No working memory (all layers self-attention)}
      & \textit{\underline{45.8}}
      & \textbf{26}
      & 85.6M \\
    \midrule
    \makecell[l]{Detach sentence reps at memory write}
      & \textit{\underline{35.0}}
      & 24
      & 85.6M \\
    \midrule
    \makecell[l]{In-context working memory}
      & 30.2
      & 19
      & 85.6M \\
    \midrule
    \makecell[l]{Sentence rep from last layer (12)}
      & 30.3
      & 21
      & 85.6M \\
    \midrule
    \makecell[l]{No context seeding (Static \texttt{<BOS>})}
      & 30.2
      & 21
      & 85.6M \\
    \midrule
    \makecell[l]{No EOS down-weighting}
      & 30.4
      & 21
      & 85.6M \\
    \midrule
    \makecell[l]{No stream curriculum (fixed $S{=}40$ sentences)}
      & 30.5
      & 21
      & 85.6M \\
    \midrule
    \makecell[l]{Max sentence length $= 32$}
      & 30.4
      & 22
      & 85.6M \\
    \bottomrule
  \end{tabular}
\end{table}

\subsection{Ablations}
\label{sec:results-ablations}

Table~\ref{tab:ablation_summary} reports ablations designed to isolate which TG components drive the gains observed in \S\ref{sec:results}. All ablation models are trained on the same $30\mathrm{M}$-token pretraining subset using the protocol in \S\ref{sec:setup}. For each run we report: (i) test perplexity (computed over lexical tokens only, matching \S\ref{sec:setup}); (ii) the number of non-embedding parameters $N$; and (iii) training throughput measured as sentence steps per second. Throughput is measured on a single NVIDIA A40 GPU with no parallelization.

\paragraph{Increasing per-layer capacity.}
The baseline TG stack alternates within-sentence self-attention and memory cross-attention layers. We evaluate higher-capacity variants in which each layer contains both attention mechanisms before the feedforward network: a serial Self$\rightarrow$Cross attention ordering and a parallel Self\&Cross ordering. Self$\rightarrow$Cross improves test perplexity from $29.8 \rightarrow 29.4$, but increases model size from $85\mathrm{M} \rightarrow 114\mathrm{M}$ non-embedding parameters ($+34\%$) and reduces throughput from $21 \rightarrow 17$ sent./sec. The Parallel variant matches the same parameter and throughput cost but yields a smaller gain (29.7 PPL), suggesting that the serial ordering is a more effective way to scale TG under this design. Overall, these results show that allocating additional parameters to TG can yield further gains and that adding per-layer cross and self attention is an effective route to scale TG.

\paragraph{Working memory and gradient flow through memory are essential.}
The largest performance drop comes from removing the working memory entirely by replacing every cross-attention block with within-sentence causal self-attention (Table~\ref{tab:ablation_summary}). This ablation effectively
reduces TG into a sentence-local decoder whose token context resets at every sentence boundary (effectively a GPT-2 with a one-sentence context window), and test perplexity collapses ($29.8 \rightarrow 45.8$). This sharp degradation indicates that TG’s performance fundamentally relies on conditioning token generation on earlier context via sentence gestalts, consistent with the broader evidence that TG increasingly up-weights and routes learning signal through the working memory pathway as training progresses (Appendices~\ref{app:memory_gate_plots} and~\ref{sec:comp_grad}). The next most consequential ablation detaches the gradients of sentence representations when updating the working memory, which considerably worsens test perplexity ($29.8 \rightarrow 35.0$); despite increasing the throughput modestly ($21 \rightarrow 24$) due to the reduced backpropagation depth. This degradation indicates that TG’s gains do not come merely from introducing a recurrent state, but from training that latent state end-to-end: next-token losses on later sentences backpropagate through memory reads to optimize the parameters that produced earlier sentence gestalts.

\paragraph{External vs.\ in-context working memory.}
Placing sentence vectors \emph{in context} as a prefix and relying on self-attention across the extended sequence (while preserving gradient flow through the memory vectors) largely retains TG’s modeling benefits (test PPL $30.2$), but is slower (throughput $19$ vs.\ $21$ sent./sec). 
This is consistent with the quadratic cost of self-attention: for token sentence length $n$ and memory size $M$, TG decouples processing into $O(n^2)$ self-attention and $O(n \cdot M)$ cross-attention, while the in-context approach self-attends to the concatenated sequence in $O((n+M)^2)$. 

\paragraph{Other design choices.}
The remaining ablations, such as the in-context memory ablation, yield modest but interpretable degradations: extracting sentence representations from the last layer, removing context seeding, disabling EOS down-weighting, disabling the stream-length curriculum, and halving the maximum sentence length all increase test perplexity by $0.4$--$0.7$ PPL (about $1$--$2.4\%$) relative to baseline. 
Among these, removing the stream curriculum produces the largest drop (PPL $30.5$), supporting the role of controlling backpropagation depth for stable optimization. 
Overall, among the components tested here, the dominant driver of performance is end-to-end training of sentence gestalts through differentiable memory, with the remaining components providing incremental improvements.

\section{Related Works}

\paragraph{Learning sentence representations.}
Many models learn explicit sentence embeddings by adding extra training objectives on top of language modeling. 
BERT, for example, adds a Next Sentence Prediction (NSP) loss, where a classifier must decide whether a second sentence truly follows the first or is just a randomly sampled sentence \citep{devlin2019bert}. 
Contrastive methods such as SimCSE \citep{gao2021simcse} and related sentence-embedding models \citep{reimers2019sentence,cer2018universal,kiros2015skip,logeswaran2018quick} instead pull together different views of the same sentence and push apart embeddings of unrelated sentences.
However, RoBERTa and a broader analysis of loss functions show that removing NSP and relying only on language modeling can improve performance, and that auxiliary sentence-level objectives are often brittle and can even hurt generalization if not tuned carefully \citep{liu2019roberta,Aroca-Ouellette2020OnLosses}.

Large Concept Model (LCM) \citep{BarraultLCM2024} similarly operates at the sentence level, but trains an autoregressive model to predict the next sentence embedding in a frozen sentence representation space (SONAR), which is then decoded into tokens, rather than directly predicting a token sequence. In contrast, the Thought Gestalt (TG) model does not use a frozen encoder or any separate sentence-level loss: token and sentence spaces are learned jointly using the same next-token prediction objective. This avoids brittle auxiliary losses while still yielding contextualized sentence embeddings that can be reused as high-level states representing prior content.

\paragraph{Sequence compression and gisting.}

Several recent approaches compress long token sequences into a small set of learned vectors.
Gisting for LLMs inserts special “gist" tokens in context after the prompt and modifies attention masks so that gist tokens attend to the full prompt, while later tokens attend only to the gists, forcing the model to compress the prompt into a few gist representations that can be cached and reused \citep{mu2023gisting}.
AutoCompressors \citep{chevalier2023autocompressors} apply a similar idea at the document level, recursively summarizing segments into short vectors that are fed back to the model, enabling long-range conditioning with only a few learned summaries.
In both of these works, summary tokens are trained only via the general next-token prediction loss, without an explicit reconstruction objective.
Compressive transformers, in contrast, build a two-tier memory of recent raw activations and older compressed activations, trained with auxiliary reconstruction losses \citep{rae2019compressivetransformer}.

In TG, sentence representations play the role of compression tokens: the model extracts a single gist vector per sentence from a mid–upper layer after full contextualization with within-sentence tokens and the sentence memory, and writes it into the working memory without detaching its computation graph. The only supervision for these sentence compression tokens comes from the same next-token losses on later sentences, backpropagated through the memory.
Unlike Compressive transformers, TG does not introduce a separate compression loss; and unlike standard gisting, the same fully contextualized sentence representation is reused in the cross-attention path at every layer (each transformer layer applies its own learned projections to these vectors).
Thus, even early layers can query a stable, contextualized summary of prior content rather than raw, noisy activations, helping alleviate long-range contextualization errors \citep{lepori2024racingthoughts}.

\paragraph{Recurrence and memory in transformers.}
Another design axis concerns extending transformers to reuse past representations to enable accessing content beyond a fixed context window. 
transformer-XL caches hidden states from previous segments and lets the current segment attend to them, providing segment-level recurrence but truncating gradients at the cached states \citep{dai2019transformerxl}.
Block-Recurrent transformers apply a transformer layer recurrently over blocks of tokens, with a recurrent state that each layer attends to, combining local self-attention within a block with an RNN-like state that carries information across blocks \citep{hutchins2022blockrecurrent}. 
Recurrent Memory transformer adds dedicated memory tokens that are passed between segments and updated by self-attention, providing a differentiable memory channel across long sequences \citep{bulatov2022rmt}.
In a complementary direction, Latent Thought Models (LTMs) introduce a memory of sequence-specific latent ``thought'' vectors inferred via iterative variational Bayes updates (inference-time computation) to guide a transformer decoder \citep{kong2025latent}.
Memorizing transformers, on the other hand, add an external key--value store of past activations that can be queried with kNN retrieval at inference time, extending the effective context while keeping the memory non-differentiable \citep{wu2022memorizing}.

TG shares with these models the design principles of reusing past computation, but its recurrent state resides in an external differentiable memory of semantically coherent sentence gestalts rather than token-level caches or memory tokens encoding arbitrary token sequences, and these gestalts are trained only via a robust language modeling objective: the next-token prediction loss.
This design ties recurrence, compression, and sentence structure together: long-range information is carried forward in compact sentence representations that are optimized end-to-end for generative performance. These semantically grounded units are natural candidates for storage and retrieval in a long-term memory system without the need to increase model parameters, in the spirit of the Memorizing transformer but using sentence- or thought-aligned chunks that are more effective than arbitrary token blocks \citep{glaforge2025sentencewindow,bhat2025rethinkingchunk}. 

\section{Conclusion}

\paragraph{Future work.}
Next steps are to scale TG along several axes and characterize emerging capabilities. This includes increasing model capacity (e.g., widen $d_{\text{model}}$ and/or add per-layer self$\rightarrow$cross capacity), extending the backpropagation dependency chain via longer sentence streams, increasing memory capacity, and training on larger, more diverse corpora. Because TG improves faster on relational-direction generalization, an interesting question is whether scaling translates into gains on reasoning and mathematical benchmarks where robust relational representations are critical. Finally, extending beyond two abstraction levels to a hierarchy of learned abstractions, paired with long-term memory for concepts at multiple granularities, are other future directions of this work to enable building situation models and continual learning.

\paragraph{Limitations.}
We study TG in a low-compute regime (up to 50M training tokens and $\sim$85M non-embedding parameters), which enables scaling sweeps and extensive ablations but is far below industry-scale LLM training. Demonstrating practical impact in real-world settings will require training larger TG models on bigger datasets, which is the next step of this work.

In conclusion, this work shows considerable promise for the use of latent thought gestalts  learned end-to-end via next-token prediction training objective. The critical next step is to scale this approach to fully realize its potential.

\bibliographystyle{unsrtnat}
\bibliography{references}

\clearpage

\appendix

\section{Recursive Backpropagation Through Working Memory}
\label{app:grad-chain}

\begin{figure}[h]
\centering
\makebox[0pt][l]{\hspace{-7.6cm}\includegraphics[width=1.08\linewidth]{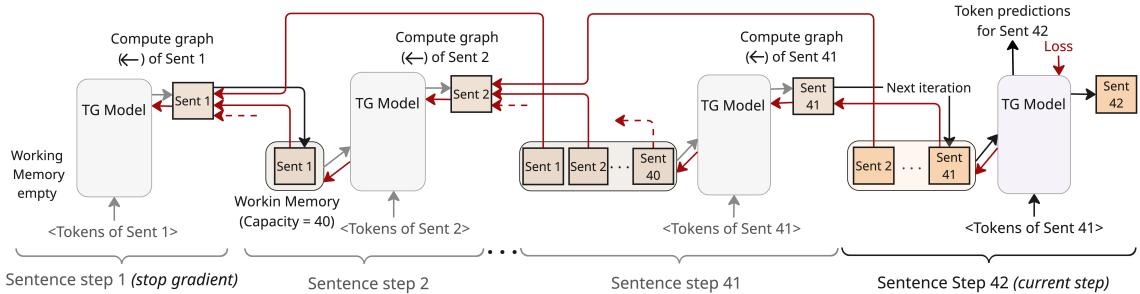}}
\caption{\small \textbf{Recursive gradient flow through working memory.}
With working-memory capacity $M{=}40$, at sentence step $t{=}42$, the forward pass cross-attends only to the current memory contents $\{\mathbf{s}_{2},\ldots,\mathbf{s}_{41}\}$ ($\mathbf{s}_{1}$ has been evicted from the fixed-capacity memory).
However, because sentence vectors are written to memory without detaching, the vectors  in the memory retain computation-graph links to the graph of earlier sentences that they attended to when they were formed. Thus, gradients from the step 42 token losses backpropagate through the stored computation graphs and reach $\mathbf{s}_{1}$ via the graph of the stored $\mathbf{s}_{2}$. The chain terminates at the first sentence of each sentence stream, where the memory is reset (stop-gradient).}
\label{fig:tg_grad}
\end{figure}

TG writes sentence vectors to working memory without detaching their computation graphs. 
While the memory size $M$ limits which sentence vectors are available to cross-attention in the forward pass, it does not limit the backward flow of gradients. The loss at step $t$ backpropagates recursively through the stored sentence vectors into older sentences that influenced the computation of the current memory entries. This recursive chain terminates at the first sentence of the stream (Sentence 1), which is computed with an empty memory. Thus, the maximum backward depth is bounded by the maximum stream length $S$ (\S\ref{sec:data-pipeline}).

\section{Other TG Design Principles}
\label{app:add_desings}

\subsection{Batch construction: Uniform Token-Budget Bucketing}
\label{app:batch-construct}

TG processes batches of document streams sentence-by-sentence, where parallelization occurs in sentence steps: at step $t$, the $t$-th sentence from every active sentence stream in the batch is stacked and run in a single forward pass in parallel. Because document lengths and by extension sentence stream lengths vary, standard random batch construction from a fixed number of examples (i.e., sentence streams) causes two critical issues: (1) \emph{GPU memory volatility}: an optimizer step is taken after processing one batch; therefore, as all computations graphs are stored within a batch for backpropagation and longer streams require storing a higher number of deeper computation graphs, GPU memory usage spikes when long streams cluster in a batch; and (2) \emph{unstable optimization}, where the number of supervised tokens fluctuates between batches, due to the variable number of lexical tokens per sentence and variable number of sentences per document, which would destabilize gradient updates.

We mitigate this with a sampling strategy for batch construction. This method involves: (i) \emph{Bucketing}: We group sentence streams by sentence count into buckets of fixed width (e.g., 5 sentences) to create groups of sentence streams with similar lengths and thus computation graph depths, allowing for shuffling and batching randomization within these buckets. (ii) \emph{Allocation}: We pre-allocate a fixed number of batches estimated by the total lexical tokens in the corpus and a given target token budget per batch. (iii) \emph{Distribution}: We iterate through buckets from longest to shortest and distribute streams into the pre-allocated batches uniformly, using a first-fit strategy. A stream is added to a batch only if the batch remains within two constraints: a target lexical token budget (ensuring consistent optimizer step magnitude) and a maximum stream count (bounding the memory overhead of maintaining parallel sentence memories during forward pass). This ensures that memory-intensive long streams are distributed uniformly rather than clustered, while every optimizer step is based on a consistent amount of training signal.

\subsection{Stochastic Regularization (Scheduled Dropouts)}
\label{app:regularization}

In TG, we apply stochastic regularization in several places. 

\textbf{Token dropout.}
    To encourage reliance on the sentence-memory pathway (rather than only short-range token context), we apply token dropout: during training, a fraction of content token embeddings within each sentence is randomly zeroed. Token dropout is warmed in over the global sentence-step counter.
    The default token-dropout rate is 0.15; this dropout is disabled for the smallest models in the parameter-scaling experiments due to instability early in training.

\textbf{Sentence-representation dropout.}
    To regularize the sentence vectors written to memory, we apply dropout to the hidden state at the \texttt{<EOS>} position before the sentence head.
    This dropout follows the same warm-in schedule as token dropout (0 / 50\% / 100\% over 2000/7000 sentence steps) with default rate 0.15.

\textbf{Attention dropout.}
    We use standard dropout on attention weights in both self-attention and cross-attention modules (default 0.2).

\textbf{Sentence-head internal dropout.}
    When the sentence head is an MLP (depth $>1$), intermediate layers include standard dropout blocks (fixed rate), following common SimCLR-style MLP regularization.

\section{Memory Gate Dynamics Over Training}
\label{app:memory_gate_plots}

Each TG cross-attention layer (i.e., the layers that read from the sentence-memory) includes a learnable scalar memory gate $g_{\mathrm{mem}}^{(\ell)}$ for $ \ell\in\{2,4,6,8,10,12\}$ that scales the memory cross-attention update before it is added through the residual path.
Thus, $g_{\mathrm{mem}}^{(\ell)}{=}1$ corresponds to an unscaled memory contribution, while $g_{\mathrm{mem}}^{(\ell)}{>}1$ amplifies the influence of the memory pathway relative to the rest of the layer’s residual computation. 
Because the self-attention residual branch is not scaled by this parameter, the gate provides a direct readout of the model’s relative weighting of sentence-memory content at each depth.

\begin{figure*}[h]
  \centering
  \begin{minipage}[t]{0.32\linewidth}
    \centering
    \includegraphics[width=\linewidth]{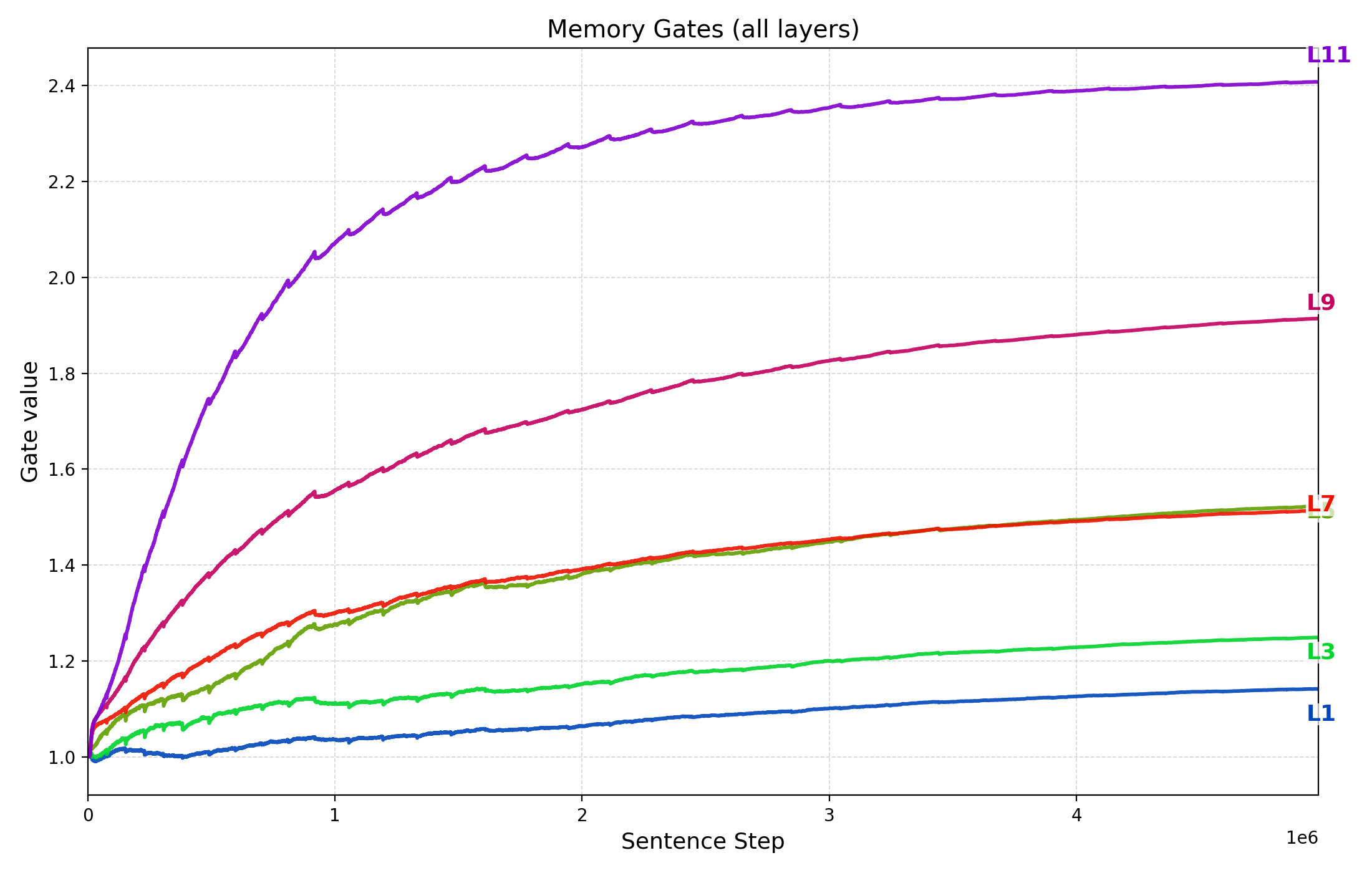}
    \vspace{-0.3em}
    {\small (a) 85M TG, 50M-token pretraining.}
  \end{minipage}\hfill
  \begin{minipage}[t]{0.32\linewidth}
    \centering
    \includegraphics[width=\linewidth]{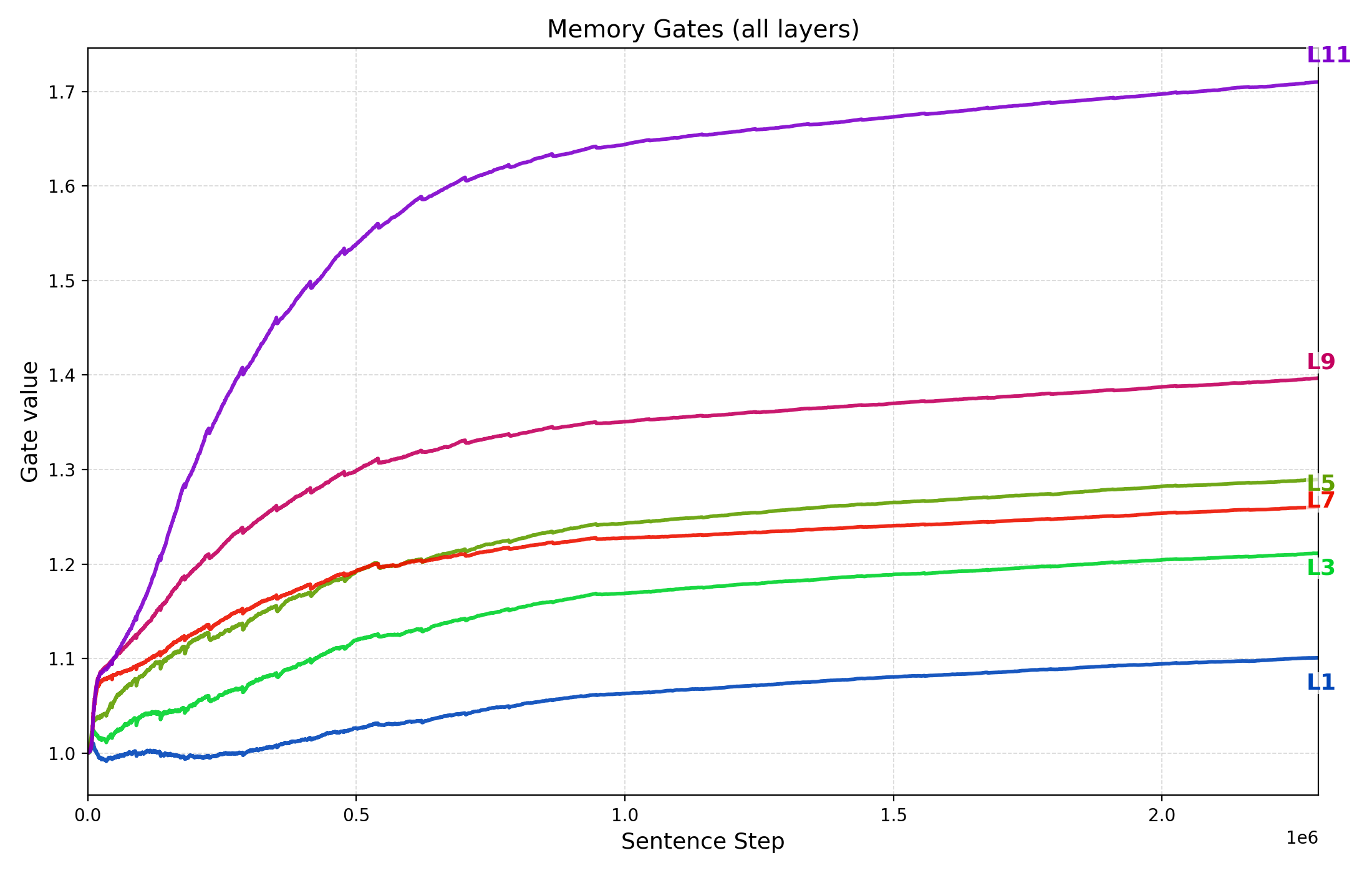}
    \vspace{-0.3em}
    {\small (b) 85M TG, 30M-token pretraining.}
  \end{minipage}\hfill
  \begin{minipage}[t]{0.32\linewidth}
    \centering
    \includegraphics[width=\linewidth]{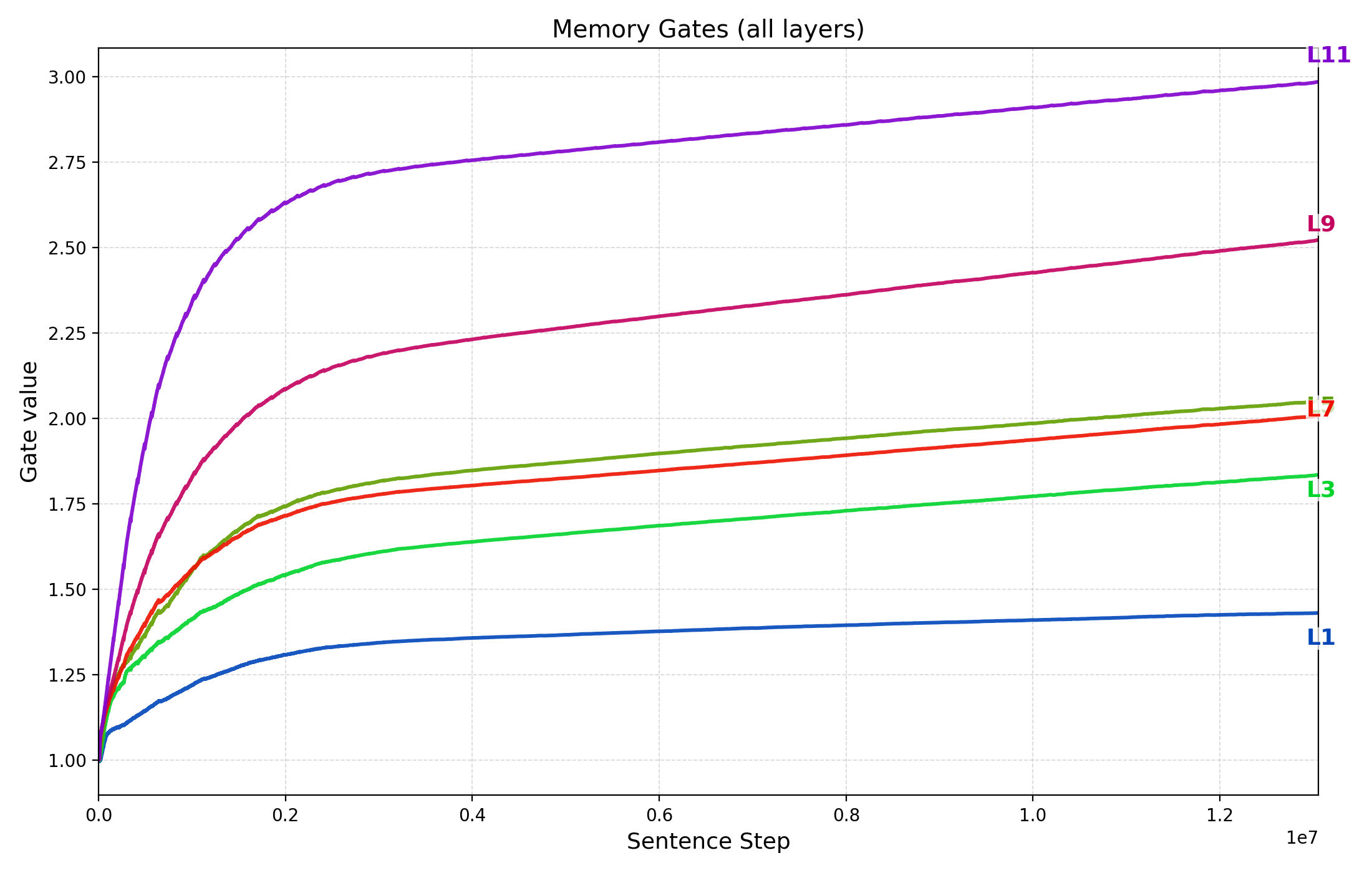}
    \vspace{-0.3em}
    {\small (c)  21M TG ($d_{\text{model}}{=}384$), 50M-token pretraining.}
  \end{minipage}

  \caption{\textbf{Learned memory-gate values over training.}
  Each curve shows the scalar gate $g_{\mathrm{mem}}^{(\ell)}$ for one cross-attention layer $\ell$ as a
  function of sentence steps. Gates grow over training and are larger in deeper layers.}
  \label{fig:memory_gates_training}
\end{figure*}

Figure~\ref{fig:memory_gates_training} plots $g_{\mathrm{mem}}^{(\ell)}$ as training progresses (x-axis is the cumulative sentence steps) up to the best validation checkpoint for three representative runs discussed in \S\ref{sec:results-scaling}: the $N{\approx}85$M baseline trained on $D{=}50$M tokens, the $N{\approx}85$M model trained on $D{=}30$M tokens, and the parameter-scaling checkpoint with $d_{\text{model}}{=}384$ ($N{\approx}21$M) trained on $D{=}50$M tokens.

Across all three settings we observe two consistent patterns: (i) gates increase early and then rise more gradually, indicating that the model learns to up-weight the memory pathway once sentence representations become informative for next-token loss; and (ii) gates are larger in higher layers (e.g., layers 9--11) than in lower layers.
This depth-wise stratification suggests higher layers rely more strongly on memory-based
information integration while lower layers remain comparatively closer to local, within-sentence
processing. The fact that gates grow steadily supports the important role of memory in next token prediction, consistent with the strong performance drop when gradient flow through memory is removed (\S\ref{sec:results-ablations}).

\section{Component Gradient Share}
\label{sec:comp_grad}

\begin{figure}[h]
\centering
\includegraphics[width=1\linewidth]{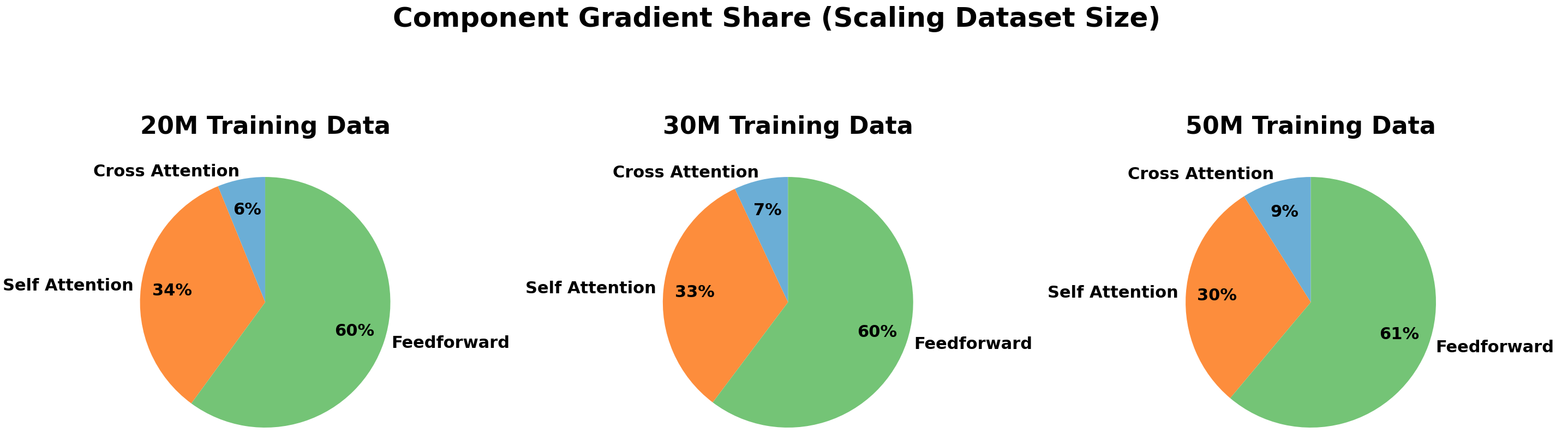}
\caption{\small \textbf{Component gradient share across dataset scaling.}
For the TG runs in the dataset-scaling experiment (\S\ref{sec:results-scaling}; $N\!\approx\!85$M non-embedding parameters),
each pie chart shows the fraction of the total next-token loss gradient magnitude attributable to parameters in
within-sentence self-attention, working memory cross-attention, and feedforward (MLP) sublayers,
averaged across layers and over training up to the early-stopping checkpoint.}
\label{fig:comp-grad}
\end{figure}

Figure~\ref{fig:comp-grad} shows a systematic shift in where the optimization signal concentrates as training data increases.
From $D{=}20$M to $30$M to $50$M tokens, the average gradient share assigned to sentence-memory cross-attention rises
(6\%$\rightarrow$7\%$\rightarrow$9\%), while the share assigned to within-sentence self-attention decreases
(34\%$\rightarrow$33\%$\rightarrow$30\%); the feedforward/MLP components remain  approximately constant
(60--61\%). We interpret the increasing cross-attention share as evidence that, at larger pretraining scales, the learned
sentence gestalts become more informative, and TG correspondingly allocates more learning capacity to the working-memory
pathway for next-token prediction. This trend is consistent with the learned memory-gate dynamics (Appendix~\ref{app:memory_gate_plots})
and with the strong degradation observed when gradient flow through memory is removed (\S\ref{sec:results-ablations}).

\section{Numerical Results}

\subsection{Scaling Efficiency Results}
\label{app:scaling_tables}
\begin{table}[H]
    \centering
    \caption{Test perplexity values corresponding to the scaling plots in Section~\ref{sec:results-scaling}. Top: Dataset scaling with fixed model size (85M non-embedding parameters). Bottom: Parameter scaling with fixed dataset size (50M tokens). Lower perplexity indicates better performance.}
    \label{tab:scaling_results}
    
    \vspace{0.5em}
    \textbf{(a) Dataset Scaling}
    \vspace{0.2em}
    
    \begin{tabular}{lrr}
        \toprule
        \textbf{Training Tokens} & \textbf{GPT-2 PPL} & \textbf{TG PPL} \\
        \midrule
        12,000,000 & 50.9 & 49.5 \\
        20,000,000 & 38.1 & 37.1 \\
        30,000,000 & 30.9 & 29.8 \\
        50,000,000 & 24.0 & 23.2 \\
        \bottomrule
    \end{tabular}

    \vspace{1em}
    \textbf{(b) Parameter Scaling}
    \vspace{0.2em}
    
    \begin{tabular}{lrr}
        \toprule
        \textbf{Params ($N$)} & \textbf{GPT-2 PPL} & \textbf{TG PPL} \\
        \midrule
        $\approx$ 340K  & 104.8 & 97.7 \\
        $\approx$ 1.3M  & 68.7  & 59.8 \\
        $\approx$ 5.4M  & 42.4  & 37.8 \\
        $\approx$ 21.3M & 28.8  & 26.8 \\
        \bottomrule
    \end{tabular}
\end{table}

\subsection{GPT-2 with Sentence Boundary Bias Results}
\label{tab:sent_bias_table}

\begin{table}[H]
    \centering
    \caption{Test perplexity (lower is better) for models trained on 20M, 30M, and 50M tokens. We compare standard GPT-2 and Thought Gestalt (TG) against a GPT-2 baseline that retains explicit sentence boundary tokens in the input stream.}
    \label{tab:baseline_sentence_boundary}
    \vspace{0.2em}
    \begin{tabular}{lccc}
        \toprule
        \textbf{Training Tokens} & \textbf{GPT-2} & \textbf{GPT-2 + Sent. Boundary} & \textbf{Standard TG} \\
        \midrule
        20,000,000 & 38.1 & 36.6 & 37.1 \\
        30,000,000 & 30.9 & 30.3 & 29.8 \\
        50,000,000 & 24.0 & 23.7 & 23.2 \\
        \bottomrule
    \end{tabular}
\end{table}

\subsection{TG with Fixed Token-Span Recurrence Results}
\label{tab:fixed_span_tables}
\begin{table}[H]
    \centering
    \caption{Test perplexity comparing the semantic sentence segmentation of Standard TG against fixed-length token chunking strategies ($N=25, 50, 75$) and the non-recurrent Standard GPT-2 baseline.}
    \label{tab:baseline_token_span}
    \vspace{0.2em}
    \begin{tabular}{lccccc}
        \toprule
        \textbf{Training Tokens} & \textbf{Standard TG} & \textbf{TG + 25-Span} & \textbf{TG + 50-Span} & \textbf{TG + 75-Span} & \textbf{GPT-2} \\
        \midrule
        20,000,000 & 37.1 & 38.9 & 38.4 & 37.8 & 38.1 \\
        30,000,000 & 29.8 & 32.0 & 31.6 & 31.5 & 30.9 \\
        50,000,000 & 23.2 & 24.7 & 24.5 & 24.2 & 24.0 \\
        \bottomrule
    \end{tabular}
\end{table}

\subsection{GPT-2 + Gist Masking Results}
\label{tab:gisting_table}

\begin{table}[H]
    \centering
    \caption{Test perplexity comparing Standard TG and GPT-2 against a GPT-2 baseline utilizing gisting-style attention masking in context \citep{mu2023gisting} to compress prior context.}
    \label{tab:baseline_gisting}
    \vspace{0.2em}
    \begin{tabular}{lccc}
        \toprule
        \textbf{Training Tokens} & \textbf{GPT-2} & \textbf{GPT-2 + Gist Masking} & \textbf{Standard TG} \\
        \midrule
        20,000,000 & 38.1 & 48.3 & 37.1 \\
        30,000,000 & 30.9 & 40.6 & 29.8 \\
        50,000,000 & 24.0 & 31.9 & 23.2 \\
        \bottomrule
    \end{tabular}
\end{table}

\end{document}